%% file: main.tex
\documentclass[letterpaper, 10pt, conference]{ieeeconf}

\IEEEoverridecommandlockouts                              
 
\let\labelindent\relax

\usepackage[loadonly]{enumitem}

\usepackage{amssymb}
\usepackage{booktabs} 
\usepackage{graphicx}
\usepackage{pifont}
\usepackage{etoolbox}
\usepackage{amsmath}
\usepackage{multirow}
\usepackage{longtable}
 \usepackage[font=footnotesize,labelformat=simple]{subcaption}
\usepackage[font=footnotesize]{caption}
\usepackage{tabularx}
\usepackage{subcaption}
\usepackage{adjustbox}
\newcommand{\cmark}
{{\color{green!60!black}\checkmark}}
\newcommand{\xmark}{{\color{red!70!black}\ding{55}}}
\newcommand{\benchmarkname}{\textsc{RoboEval}}
\usepackage{ltxtable}
\usepackage{wrapfig}
\usepackage{array, rotating, longtable}
\usepackage[table,dvipsnames]{xcolor}

\usepackage{makecell}

\newcolumntype{C}{>{\centering\arraybackslash}X}

\usepackage{listings}
\definecolor{codegreen}{rgb}{0,0.6,0}
\definecolor{codegray}{rgb}{0.5,0.5,0.5}
\definecolor{codepurple}{rgb}{0.58,0,0.82}
\definecolor{backcolour}{rgb}{0.95,0.95,0.92}
\lstdefinestyle{mystyle}{
    backgroundcolor=\color{backcolour},   
    commentstyle=\color{codegreen},
    keywordstyle=\color{magenta},
    numberstyle=\tiny\color{codegray},
    stringstyle=\color{codepurple},
    basicstyle=\ttfamily\footnotesize,
    breakatwhitespace=false,         
    breaklines=true,                 
    captionpos=b,                    
    keepspaces=true,                 
    numbers=left,                    
    numbersep=5pt,                  
    showspaces=false,                
    showstringspaces=false,
    showtabs=false,                  
    tabsize=2
}
\makeatletter
\renewcommand{\@seccntformat}[1]{%
  \ifx#1\subsection \thesubsection.\quad
  \else \csname the#1\endcsname.\quad \fi}
\makeatother

\usepackage{hyperref}

\title{\LARGE \bf {\includegraphics[width=3.1cm]{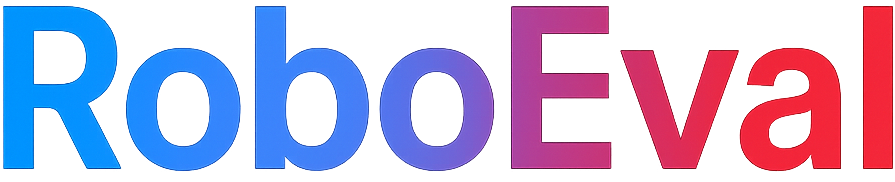}}: Where Robotic Manipulation Meets Structured and Scalable Evaluation}

\author{%
\begin{tabular}{c}
Yi Ru Wang$^{1,2\dagger}$, Carter Ung$^{1}$, Christopher Tan$^{1}$,
Grant Tannert$^{1}$, Jiafei Duan$^{1,2}$, Josephine Li$^{1}$, Anh Le$^{1}$\\
Rishabh Oswal$^{1}$, Markus Grotz$^{1}$, Wilbert Pumacay$^{1,2}$,
Yuquan Deng$^{1}$, Ranjay Krishna$^{1,2}$\\
Dieter Fox$^{1,2*}$, Siddhartha Srinivasa$^{1*}$
\end{tabular}%
\thanks{$^{1}$University of Washington, Seattle, WA, USA}%
\thanks{$^{2}$Allen Institute for Artificial Intelligence (AI2), Seattle, WA, USA}%
\thanks{$^{\dagger}$Corresponding author: yiruwang@cs.washington.edu}%
\thanks{$^{*}$Dieter Fox and Siddhartha Srinivasa contributed equally as advisors.}%
}

\begin{document}

\maketitle

\begin{figure*}[htbp]
    \centering
    \includegraphics[width=\linewidth]{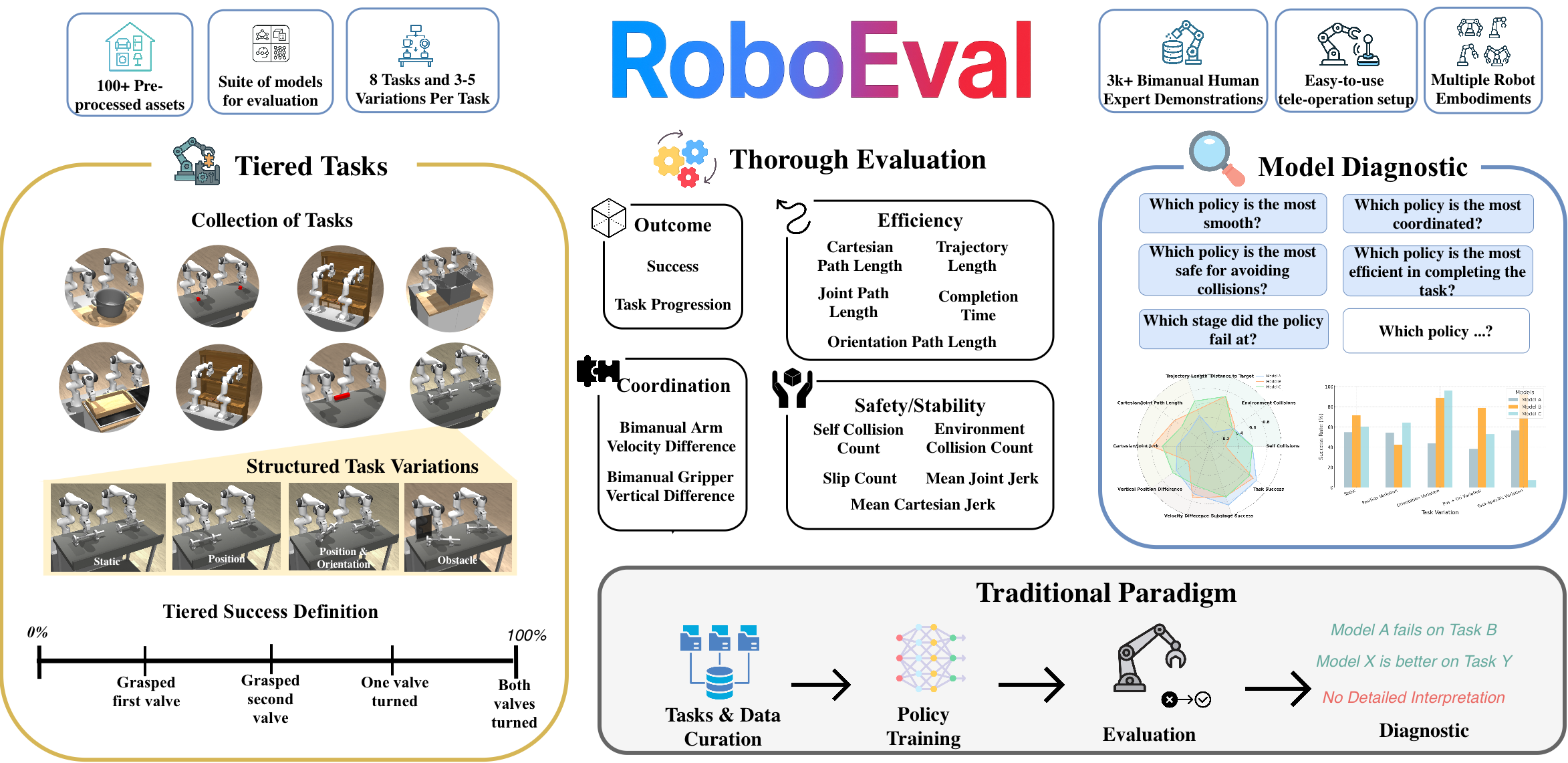} 
    \caption{
    \textbf{Overview of \benchmarkname.} 
    \benchmarkname~provides a principled evaluation framework that goes beyond binary success by integrating a rich set of \textit{behavioral and outcome metrics}. These capture execution quality in terms of trajectory fluency, spatial precision, and bimanual coordination, as well as outcome measures of task progression and structured failures. 
    The benchmark features 3,000 human-collected demonstrations across 8 tasks with 3–5 systematic variations each, a standardized asset library for constructing new tasks and perturbations, and a VR-based teleoperation interface for realistic data collection. Together, these components enable fine-grained policy diagnosis and reproducible evaluation.
}

\label{fig:system}
\vspace{-1em}
\end{figure*}

\thispagestyle{empty}
\pagestyle{empty}

\begin{abstract}

We introduce \textbf{\benchmarkname}, a structured evaluation framework and benchmark for robotic manipulation that augments binary success with principled behavioral and outcome metrics. Existing evaluations often collapse performance into outcome counts, masking differences in execution quality and obscuring failure structure. \benchmarkname~provides eight bimanual tasks with systematically controlled variations, more than three thousand expert demonstrations, and a modular simulation platform for reproducible experimentation. All tasks are instrumented with standardized metrics that quantify efficiency, coordination, and safety/stability, as well as outcome measures that trace stagewise progress and localize failure modes. Through extensive experiments with state-of-the-art visuomotor policies, we validate these metrics by analyzing their stability under variation, discriminative power across policies with similar success rates, and correlation with task success. Project Page: \href{https://robo-eval.github.io}{https://robo-eval.github.io}

\end{abstract}

\input{sections/1_introduction}
\input{sections/2_related_works}

\input{sections/3_method}

\input{sections/4_experiments}
\input{sections/6_discussion_conclusion}

\addtolength{\textheight}{-12cm}  


\section*{Acknowledgments}
\small Yi Ru Wang is supported by the Natural Sciences and Engineering Research Council of Canada Postgraduate Scholarships – Doctoral program (NSERC-PGSD). This work was partially supported by the National Science Foundation NRI program (\#2132848), DARPA RACER (\#HR0011-21-C-0171), the Office of Naval Research (\#N00014-24-S-B001 and \#2022-016-01 UW), and the DEVCOM Army Research Laboratory (Award: W911NF-24-2-0191). We gratefully acknowledge support from Amazon and the Allen Institute for Artificial Intelligence (AI2), as well as gifts from Collaborative Robotics, Cruise, and other industry partners.



{\footnotesize
\bibliographystyle{IEEEtran}
\bibliography{example}
}

\end{document}

%% file: sections/1_introduction.tex
\section{Introduction}

Robotic manipulation is most often evaluated by binary success \cite{james2020rlbench,chernyadev2024bigymdemodrivenmobilebimanual,jiang2024dexmimicen,peract2,robotwin}. Yet the true capabilities of a policy are reflected in its motions: the stability of a grasp, the smoothness of a trajectory, and the coordination between arms. Two policies may achieve the same success rate, one fluent and robust and the other brittle and erratic, but conventional evaluation renders them indistinguishable. This limitation becomes more pronounced as policies operate in increasingly complex environments. Without richer measures, we cannot explain how policies behave, why they fail, or what skills they possess. \textit{In this work, we challenge the assumption that binary success adequately reflects policy performance, as policies may succeed while concealing critical failures beyond outcome-only evaluation.}

Behavioral metrics address this gap by quantifying execution quality rather than outcomes alone, including trajectory smoothness, contact stability, and coordination timing. Other domains have benefited from similar shifts toward richer evaluation signals, such as BLEU in language \cite{papineni2002bleu}, FID in vision \cite{heusel2017gans}, and Elo ratings in games, which measure how well systems perform rather than only whether they succeed.
The use of multi-dimensional metrics is well grounded in prior robotics literature. Classical standards and benchmarking efforts characterize performance along axes such as accuracy, efficiency, stability, and safety \cite{kimble2022performance,falco2015grasping,madhavan2009benchmarking,bostelman2016survey}. Motivated by this perspective, we organize behavioral metrics into three axes: \emph{efficiency}, capturing time and resource usage; \emph{safety/stability}, capturing robustness and avoidance of undesirable events such as collisions or slips; and \emph{coordination}, capturing temporal and spatial coupling between actions. Complementing these, outcome-driven metrics measure \emph{task progression} and \emph{binary task success}, enabling analysis of both how a task is completed and whether it is completed.

A central question is whether behavioral metrics provide reliable and meaningful insight into policy performance. To be useful, they should be \emph{skill-relevant}, reflecting capabilities that drive success; \emph{discriminative}, distinguishing policies with similar success rates but different execution quality; and \emph{consistent}, remaining informative under variation in task configuration and scene complexity.

We introduce \textbf{\benchmarkname}, an evaluation framework for studying behavioral metrics in robotic manipulation. It includes eight bimanual tasks with controlled variations, over three thousand expert demonstrations collected via VR teleoperation, and a modular simulation platform for reproducible experimentation. Each task is instrumented with behavioral metrics across efficiency, safety or stability, and coordination, along with outcome metrics capturing task progression and binary success. Using this framework, we examine whether these metrics satisfy the proposed criteria by analyzing their stability under task variation, their ability to distinguish policies with similar success rates, and their relationship to task success. Beyond this study, \benchmarkname\ serves as a platform for fine-grained evaluation and extension to new tasks and metrics. Through experiments with state-of-the-art visuomotor policies, we show that \benchmarkname\ provides insights beyond success rates. Behavioral metrics reveal which aspects of execution contribute to robustness and distinguish policies with similar success but different execution quality. Outcome metrics expose structured failure modes, enabling more precise diagnosis of policy limitations. Task variations highlight when metrics remain informative and when success alone becomes insufficient.

\textbf{Contributions.} This work makes three contributions: (1) an evaluation framework that augments binary success with behavioral and outcome metrics for fine-grained analysis, (2) a validation study demonstrating that the metrics are stable, discriminative, and complementary to task success, and (3) a reusable benchmark with eight tasks, variations, and 3,000 demonstrations, supporting extensible research.

%% file: sections/2_related_works.tex
\section{Related Works}
\vspace{-0.1em}

\textbf{Benchmarks for Robotic Manipulation.} Significant progress has been made in benchmarking single-arm manipulation \cite{robosuite2020, gu2023maniskill2, james2020rlbench,pumacay2024colosseumbenchmarkevaluatinggeneralization,li2024evaluating}. Recent efforts such as Peract2 \cite{peract2}, BiGym \cite{chernyadev2024bigymdemodrivenmobilebimanual}, and RoboTwin \cite{robotwin} extend this to scalable and data-rich bimanual settings, using scripted, VR-collected, and synthetic data respectively. HumanoidBench \cite{humanoidbench2024} further broadens evaluation to whole-body humanoid manipulation, while other work studies dexterous hand use \cite{chen2023bi}. Despite these advances, existing benchmarks provide limited insight into when and why policies fail, particularly in coordinated bimanual settings. Real-world evaluation efforts are often restricted to single tasks \cite{luo2024benchmarking}, limiting diversity. Our work complements prior benchmarks by enabling structured, fine-grained evaluation and unifying diverse tasks within a common framework (Table \ref{tab:benchmark-comparison}).

\textbf{Evaluation Metrics for Manipulation.}
Evaluation metrics are central to assessing manipulation performance as tasks grow in complexity. Prior work surveys common metrics such as success rate, accuracy, and robustness \cite{newbury2023deep}, while related domains emphasize the importance of task-specific, fine-grained evaluation \cite{anderson2018evaluation}. Benchmarks such as RB2 \cite{dasari2022rb2roboticmanipulationbenchmarking} and Colosseum \cite{pumacay2024colosseumbenchmarkevaluatinggeneralization} evaluate generalization under environmental variation. However, existing approaches remain fragmented, focusing on isolated measures of success or robustness without a unified framework that captures task progression, execution quality, and failure modes.

%% file: sections/3_method.tex
\section{\benchmarkname~Benchmark}

\begin{figure*}[htbp]
    \centering
    \includegraphics[width=0.8\linewidth]{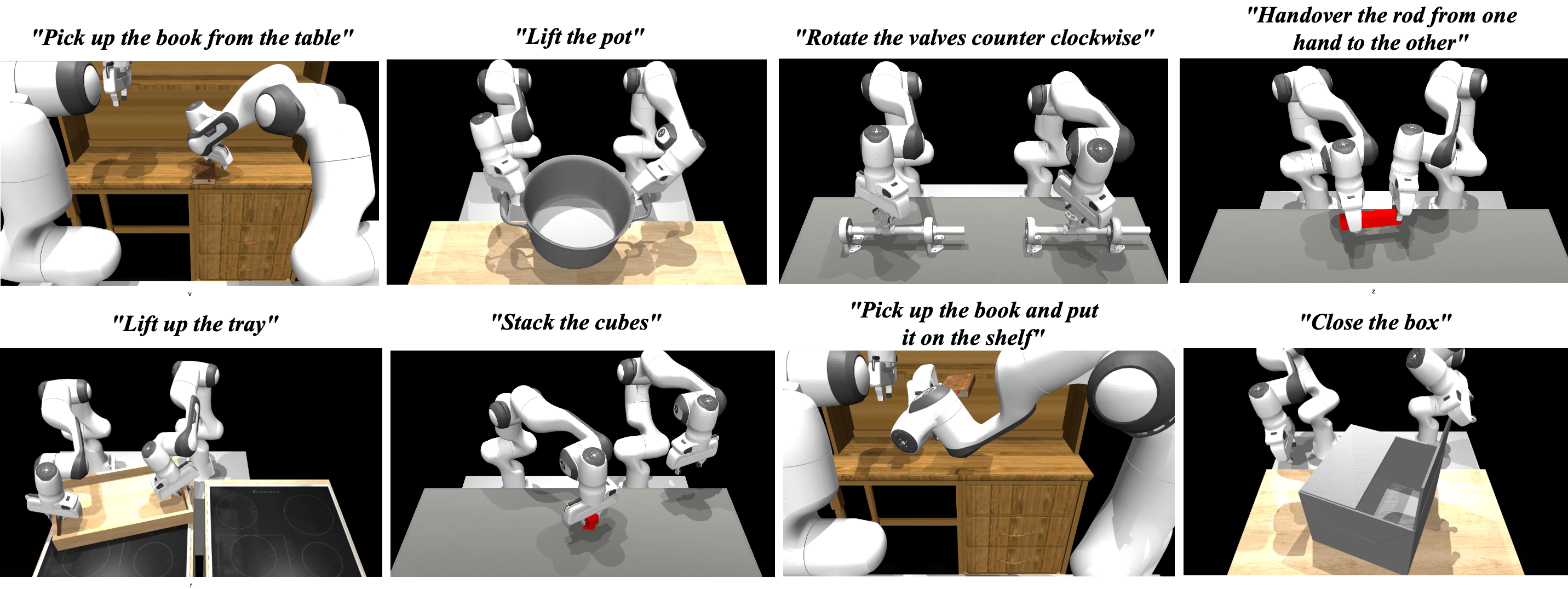} 
    \caption{\textbf{Base tasks in \benchmarkname.} \benchmarkname~introduces an initial suite of 8 bimanual manipulation tasks, each accompanied by 3–4 structured variations. All tasks are instrumented with behavior metric logging and task-stage definitions to support fine-grained progress and outcome analysis. The benchmark is modular by design, allowing for seamless integration of new tasks to accommodate evolving research needs within the community.}
    \label{fig:your_label}
    \vspace{-1em}
\end{figure*}

\benchmarkname~is designed to evaluate bimanual manipulation policies under controlled yet diverse conditions, enabling principled analysis of both behavioral and outcome metrics. The first release includes 8 base tasks spanning service, warehouse, and industrial domains, such as lifting a tray, closing a box, and rotating a hand-wheel. Each task is paired with systematic variations in spatial configuration, from fixed layouts to randomized object poses, creating a controlled testbed for analyzing metric stability, discriminative power, and sensitivity to task complexity.
To ground evaluation in real behavior, we provide 3000 human demonstrations across all tasks and variations. These support imitation learning and serve as baselines for evaluating policy fluency and precision. All environments are instrumented with fine-grained behavioral metrics across efficiency, safety/stability, and coordination, alongside outcome metrics that capture task progression and overall success.
Together, these components form a modular and extensible simulation framework for reproducible, data-driven evaluation. An overview of \benchmarkname~is shown in Figure \ref{fig:system}. The remainder of this section describes the design philosophy (Section \ref{subsection:design_philosophy}), base tasks (Section \ref{subsection:tasks}), dataset statistics (Section \ref{subsection:stats}), and evaluation protocol (Section \ref{subsection:evaluation}).

\input{tables/benchmark_comparison}

\subsection{Design Philosophy}
\label{subsection:design_philosophy}
\benchmarkname~is designed as a comprehensive benchmark for evaluating learning-based manipulation policies, grounded in three principles. \textit{Diversity:} Real-world bimanual manipulation spans a wide range of task structures, object geometries, and control challenges. \benchmarkname~captures this through tasks with varying temporal complexity, coordination demands, and semantic content, from non-prehensile pushing to tightly coupled bimanual behaviors. This ensures evaluation reflects both individual skills and their generalization across tasks. \textit{Interpretability:} Binary success provides limited insight into policy behavior. \benchmarkname~enables structured, fine-grained analysis through multi-dimensional behavioral and outcome metrics, allowing diagnosis of failure modes, sensitivity to variation, and qualitative differences between policies. \textit{Extensibility:} \benchmarkname~separates task definitions, variation generation, and evaluation protocols through standardized interfaces. New tasks, variations, and metrics can be added modularly, enabling extension while preserving comparability across studies.

\subsection{Tasks}
\label{subsection:tasks}
Tasks in \benchmarkname~span diverse settings and skill requirements, forming a systematic testbed for evaluating robotic manipulation. Each task is a goal-conditioned episode with explicit success criteria reflecting common household, industrial, and tabletop interactions. The suite covers both short-horizon objectives (e.g., lifting an object) and long-horizon, multi-step activities (e.g., cleaning a desk by placing a book on a shelf), with stage-wise progress tracking to capture partial success.

\textbf{Task Definition.} Each task in \benchmarkname~is defined by the tuple $\mathcal{T} = (\mathcal{S}, \mathcal{A}, \mathcal{P}, \mathcal{G}, \rho_0, \mathcal{S}_\text{success})$. The state space $\mathcal{S}$ includes robot joint states, object poses, and environmental context; the action space $\mathcal{A}$ consists of continuous control inputs such as joint positions and end-effector delta displacements; and $\mathcal{P}$ denotes the transition dynamics governed by a physics simulator. The goal space $\mathcal{G}$ specifies the intended outcome of the task, while the success set $\mathcal{S}_\text{success} \subset \mathcal{S}$ defines binary completion based on thresholded geometric conditions (e.g., object pose alignment or contact). Tasks are initialized by sampling from an initial state distribution $\rho_0$. Agents in \benchmarkname~learn from a dataset of expert demonstrations $\mathcal{D}_\mathcal{T} = \{(s_0, a_0, \ldots, s_T)\}$, collected via human teleoperation. To support fine-grained analysis, each task is instantiated with a parameterized family of variants $\mathcal{T}_\theta$ where $\theta \in \Theta$ modulates scene layout or semantic content.

\textbf{Skill Diversity.} Table~\ref{tab:task-set} summarizes the range of tasks included in the benchmark. Drawing on the bimanual taxonomy in~\cite{bimanual_taxonomy}, the suite is designed to span diverse motor and coordination demands, from single-arm actions such as turning a valve, to long-horizon interactions like packing a box, to tightly coordinated dual-arm motions such as lifting and balancing a tray. This diversity ensures that the benchmark probes capabilities in coordination, precision, and smooth execution across different task requirements.

\textbf{Task Variations.} 
The initial release of \benchmarkname~includes a curated set of bimanual manipulation tasks with structured spatial variations designed to probe robustness and generalization. 
These variations challenge visuomotor policies to adapt their coordination strategies while preserving task semantics. Future extensions of \benchmarkname~can build on this foundation by incorporating additional modalities, such as visual distractors, lighting changes, and variations in object physical properties.

\textbf{Task Design.} 
\benchmarkname~includes a modular task generation pipeline that allows new or external tasks to be authored and integrated with minimal code overhead. A unified interface and built-in behavioral and outcome metrics enable principled evaluation while supporting the development of generalist manipulation policies. As shown in Table~\ref{tab:benchmark-comparison}, \benchmarkname~distinguishes itself from existing benchmarks through tiered task variations, fine-grained metrics, and a large-scale repository of expert human demonstrations, creating a comprehensive platform for benchmarking bimanual manipulation.

\subsection{Task and Dataset Statistics}
\label{subsection:stats}

In total, \benchmarkname~introduces 3,000 high-quality human expert demonstrations for bimanual manipulation, making it one of the largest collections of natural teleoperated bimanual demonstrations. These demonstrations were collected using a VR-based teleoperation system, enabling precise and dexterous control over dual-arm manipulators in diverse scenarios. Table~\ref{tab:task-set} provides a breakdown of the task categories, associated variation schemes, and the number of demonstrations per task. The initial task suite in \benchmarkname~spans 8 core manipulation skills, including prehensile actions such as grasping and lifting, as well as non-prehensile strategies like pushing. These tasks are further characterized by varying amount of spatial complexities. Due to the natural variability inherent in human demonstrations, the dataset exhibits significant diversity in execution strategies, motion trajectories, and coordination styles. This variability is critical for robust learning and generalization. Importantly, \benchmarkname~not only offers scale, but also supports fine-grained analysis by capturing rich multimodal signals. We capture proprioception, visual observations, and scene-annotated interaction states, enabling detailed diagnostics of policy behavior across the spatial, temporal, and coordination axes. 

\subsection{Evaluation Scoring}
\label{subsection:evaluation}

We introduce four classes of metrics to systematically evaluate policy performance, encompassing both behavioral quality and task-level outcomes. Behavioral metrics are grouped into three axes: \emph{efficiency}, \emph{safety/stability}, and \emph{coordination}. Outcome-driven metrics include \emph{task progression} and \emph{binary task success}.

\textbf{Efficiency-Based Metrics.}  We measure efficiency using both temporal and spatial properties of the trajectory. For temporal efficiency, we track the \emph{trajectory length} (number of control steps) and \emph{completion time} (total duration in seconds).  For spatial efficiency, we measure using trajectory length in joint, Cartesian, and orientation spaces. Specifically, we compute \emph{joint path length} and \emph{Cartesian path length} as the cumulative displacement along the trajectory:
\begin{equation}
\mathcal{L}_{\text{joint}} = \sum_{t=1}^{T-1} \| q_{t+1} - q_t \|_2, \quad
\mathcal{L}_{\text{cart}} = \sum_{t=1}^{T-1} \| x_{t+1} - x_t \|_2,
\end{equation}
where \(q_t\) denotes the joint configuration and \(x_t\) the Cartesian end-effector position at timestep \(t\).

\textbf{Safety/Stability Metrics.}  
We evaluate physical interaction quality and execution stability through both motion-based and contact-based indicators. For motion smoothness, we compute \emph{joint jerk} and \emph{Cartesian jerk}, defined as the average norm of the third-order finite difference of the trajectory, normalized by the control timestep \(\Delta t\):
\begingroup
\scriptsize
\begin{equation}
\text{Jerk}_{\text{joint}} = \frac{1}{T-3} \sum_{t=1}^{T-3} \left\| \frac{q_{t+3} - 3q_{t+2} + 3q_{t+1} - q_t}{(\Delta t)^3} \right\|_2
\end{equation}
\begin{equation}
\text{Jerk}_{\text{cart}} = \frac{1}{T-3} \sum_{t=1}^{T-3} \left\| \frac{x_{t+3} - 3x_{t+2} + 3x_{t+1} - x_t}{(\Delta t)^3} \right\|_2.
\end{equation}
\endgroup

We additionally track three contact-based indicators: the number of \emph{self-collisions} (contacts between robot links), \emph{environment collisions} (contacts with scene elements such as tables or walls), and \emph{object slips} (instances where a grasped object unintentionally loses contact with the gripper). Together, these metrics capture motion smoothness, contact stability, and overall control reliability. High values typically indicate unstable execution or imprecise interactions.

\textbf{Coordination and Bimanual Metrics.}  
Effective bimanual manipulation requires both spatial alignment and temporal synchronization between arms. Let \(x_t^{(L)}, x_t^{(R)} \in \mathbb{R}^3\) denote the Cartesian positions of the left and right end-effectors at timestep \(t\), and \(\Delta t\) the control interval.

\textit{(1) Height Discrepancy.}  
We compute the mean absolute difference in the vertical (z-axis) positions:
\begin{equation}
\Delta z = \frac{1}{T} \sum_{t=1}^{T} \left| x_t^{(L)}[z] - x_t^{(R)}[z] \right|.
\end{equation}

\textit{(2) Velocity Divergence.}  
Let \(v_t^{(L)} = \frac{x_{t+1}^{(L)} - x_t^{(L)}}{\Delta t}\) and \(v_t^{(R)} = \frac{x_{t+1}^{(R)} - x_t^{(R)}}{\Delta t}\). We define:
\begin{equation}
\Delta v = \frac{1}{T-1} \sum_{t=1}^{T-1} \left\| v_t^{(L)} - v_t^{(R)} \right\|_2.
\end{equation}

Lower values of \(\Delta z\) and \(\Delta v\) indicate better spatial and temporal coordination, respectively.

\textbf{Task Progression and Outcome Metrics.}  
We log \emph{stage-wise success indicators} as binary flags corresponding to discrete phases of the task. Overall task success is measured as the proportion of episodes that achieve completion across evaluation rollouts.

%% file: tables/benchmark_comparison.tex
\begin{table}[t!]
\centering
\caption{\textbf{Benchmark Comparison.} We compare six manipulation benchmarks across task design, evaluation, and data. \textbf{\benchmarkname} uniquely integrates tiered bimanual tasks, structured variations, behavior metrics, task progression tracking, and human demonstrations.}
\label{tab:benchmark-comparison}
\resizebox{\linewidth}{!}{%
\begin{tabular}{lccccccc}
\toprule
\textbf{Benchmark} & \multicolumn{2}{c}{\textbf{Task Features}} & \multicolumn{3}{c}{\textbf{Evaluation Features}} & \multicolumn{2}{c}{\textbf{Data Features}} \\
\cmidrule(lr){2-3} \cmidrule(lr){4-6} \cmidrule(lr){7-8}
 & \textbf{Tiered} & \makecell[c]{\textbf{Structured}\\\textbf{Variations}} & \textbf{Success} & \makecell[c]{\textbf{Behavior}\\\textbf{Metrics}} & \makecell[c]{\textbf{Task Prog.}\\\textbf{Metrics}} & \makecell[c]{\textbf{Demo-}\\\textbf{Driven}} & \makecell[c]{\textbf{\# Expert}\\\textbf{Human Demos}} \\
\midrule
RLBench          & \xmark & \xmark & \cmark & \xmark & \xmark & \cmark & 0 \\
Bigym            & \xmark & \xmark & \cmark & \xmark & \xmark & \cmark & 2k \\
DexMimicGen      & \xmark & \xmark & \cmark & \xmark & \xmark & \cmark & 400 \\
PerAct2          & \xmark & \xmark & \cmark & \xmark & \xmark & \cmark & 0 \\
HumanoidBench    & \xmark & \xmark & \cmark & \xmark & \xmark & \xmark & 0 \\
RoboTwin         & \xmark & \xmark & \cmark & \xmark & \xmark & \cmark & 300 \\
\textbf{\benchmarkname~(Ours)} & \cmark & \cmark & \cmark & \cmark & \cmark & \cmark & 3k+ \\
\bottomrule
\end{tabular}%
}
\vspace{-0.5em}
\end{table}

\begin{table}[t!]
\centering
\caption{\textbf{Base Task Set in \benchmarkname.} We summarize the base tasks with their variation types and demonstration statistics. Variation types include static setups and spatial perturbations in position (Pos), rotation (Rot), and combined (PR).}
\label{tab:task-set}
\resizebox{0.8\linewidth}{!}{%
\begin{tabular}{lccc}
\toprule
\textbf{Task Name} & \textbf{Variations} & \textbf{\# Demos} & \textbf{Traj Len} \\
\midrule
Lift Tray                 & Static, Pos, Rot, PR      & 543 & 67.584 \\
Stack Two Cubes           & Static, Pos, Rot, PR            & 492 & 107.047 \\
Stack Single Book Shelf   & Static, Pos, PR                 & 202 & 172.302 \\
Rod Handover              & Static, Pos, Rot, PR  & 408 & 93.529 \\
Lift Pot                  & Static, Pos, Rot, PR            & 176 & 53.170 \\
Pack Box                  & Static, Pos, Rot, PR            & 394 & 133.216 \\
Pick Book From Table      & Static, Pos, Rot, PR            & 366 & 105.984 \\
Rotate Valve              & Static, Pos, PR            & 349 & 119.610 \\
\bottomrule
\end{tabular}%
}
\vspace{-2em}
\end{table}

%% file: sections/4_experiments.tex
\section{Results}

The goal of our experiments is to evaluate not only how policies perform on the benchmark, but also what can be learned from the proposed metrics themselves. To this end, we organize the study around three guiding research questions:  

\begin{itemize}
    \item \textbf{RQ1:} Do behavioral and outcome metrics provide insights beyond binary success rates (Section \ref{sec:rq1})?  
    \item \textbf{RQ2:} Under what task or policy conditions are these metrics most informative (Section \ref{sec:rq2})?  
    \item \textbf{RQ3:} How do their patterns and discriminative ability vary across domains and levels of complexity (Section \ref{sec:rq3})?  
\end{itemize}

Addressing these questions requires evaluating representative policies across a diverse set of bimanual manipulation tasks and systematically varied configurations, allowing us to examine both overall performance and the diagnostic value of different metrics.

\subsection{Implementation Details}
\textbf{Models.} We evaluate five policy architectures: ACT~\cite{zhao2023learning}, Diffusion Policy~\cite{chi2023diffusionpolicy}, GR00T N1.6~\cite{bjorck2025gr00t}, X-VLA~\cite{zheng2025x}, and $\pi_{0.5}$~\cite{pi05}. ACT and Diffusion Policy use official LeRobot implementations with ResNet-18 encoders and a 16-step action horizon, trained as single-task specialists. GR00T N1.6 (3B) is a vision-language-action model, where we fine-tune the projector and action head from the pretrained checkpoint. X-VLA (0.9B) is a soft-prompted VLA with a Florence2 backbone and flow-matching policy head. $\pi_{0.5}$ combines SigLIP, Gemma-2B, and a Gemma-300M action head, fine-tuned from a pretrained checkpoint in both full and LoRA settings. All models use AdamW with consistent hyperparameters unless noted. We also include human teleoperation demonstrations as an upper-bound reference.

\textbf{Tasks and Variations.} Our main experiments focus on 8 tasks (Table~\ref{tab:task-set}), each with 3-4 spatial variations. \textit{Static} denotes minimal scene changes, while \textit{position} and \textit{orientation} introduce spatial perturbations.

\subsection{RQ1: To what extent do behavioral and outcome metrics provide insights beyond binary success rates?}
\input{tables/task_performance_main}

\label{sec:rq1}
\textbf{Metrics offer complementary insights beyond binary success rates.} Table~\ref{tab:metrics_summary_combined} compares policies across success, task progression, and behavioral metrics on combined variations. Across eight tasks, the top-success model leads in only 34 of 104 other metric comparisons (33\%), showing that success is a poor proxy for overall quality. This gap is especially clear in smoothness: ACT achieves the lowest Cartesian and joint jerk in all tasks, up to $3\times$ lower than the next-best policy in \texttt{Rotate Valve} ($1.0$ vs.\ $3.0$\,m/s$^3$), despite leading in success on only one task. Similarly, in \texttt{Pack Box}, Pi-0.5 and ACT have comparable success ($0.54$ vs.\ $0.48$), yet ACT achieves $4\times$ lower jerk and $2.7\times$ shorter paths, revealing efficiency differences not captured by success. No single model dominates safety: collisions, self-collisions, and slips vary by task across ACT, Pi-0.5, and Diffusion. Task progression further distinguishes policies when success fails: in \texttt{Lift Tray}, three policies have $0\%$ success but reach $35$--$48\%$ progression, and in \texttt{Pick Single Book From Table}, $0\%$-success models range from $2\%$ to $22\%$. Safety metrics also expose distinct failure modes: in \texttt{Stack Two Blocks}, GR00T and Diffusion have similar success ($4\%$, $6\%$), but GR00T incurs far more self-collisions (23.6 vs.\ 2.0 per episode). Together, these results show that behavioral and outcome metrics provide a much richer view of policy competence than success alone.

\input{figures/significance_matrix}

\textbf{Behavioral metrics cluster into interpretable families that capture
distinct aspects of policy behavior.}
Figure~\ref{fig:metric_dendrogram}(a) reports pairwise Spearman correlations
across all task--variant bins, and
Figure~\ref{fig:metric_dendrogram}(b) presents a hierarchical clustering using
correlation distance ($1 - |\rho|$).
Within each evaluation-intent group, tightly correlated subsets emerge:
path-length metrics (CPL, JPL, OPL; $\rho \geq 0.95$),
jerk metrics (MCJ, MJJ; $\rho = 0.96$),
and temporal metrics (CT, TL; $\rho = 0.94$) each merge below
distance $0.07$ in the dendrogram.
However, the broader tree does not collapse into a single branch.
The temporal metrics cluster with the outcome measures (Success, TP)
at distance $0.42$, despite a \emph{negative} correlation
($\rho \approx {-}0.6$), before joining the path-length group at $0.54$,
indicating that completion time relates more closely to task success
than spatial path efficiency does.
The two coordination metrics fall into separate branches:
BAVD groups with the smoothness metrics at distance $0.51$, while
BGVD groups with path-length and collision metrics at $0.37$,
despite sharing $\rho = 0.55$, suggesting they capture complementary
behavioral dimensions.
Overall, the full metric suite spans multiple distinct branches,
supporting the use of diverse metric families to evaluate policy
quality beyond binary success.

\textbf{Behavioral metrics benchmark policies against human demonstrations.} The human row in Table~\ref{tab:metrics_summary_combined} provides a reference for efficient and reliable execution from teleoperated demonstrations. Policies consistently underperform humans on efficiency: none match human completion time or trajectory lengths, with the fastest policies $3$--$10\times$ slower and trajectories $2$--$9\times$ longer. Humans also exhibit near-zero self-collisions and slips, while all policies incur higher counts, indicating lower reliability. In contrast, policies often appear smoother: ACT achieves lower Cartesian jerk than humans in all tasks, and Pi-0.5 in six of eight, with values as low as $1.0$\,m/s$^3$ versus $3.6$\,m/s$^3$ for humans. Overall, policies have lower Cartesian jerk in 26 of 48 comparisons and lower joint jerk in 21 of 48. This reflects smoothing from learned action generation rather than better control, as teleoperation introduces discrete corrections that increase jerk. These comparisons show that behavioral metrics reveal where policies approach or exceed human execution and where clear gaps remain, which success alone cannot capture.

\subsection{RQ2: Under what conditions are the metrics most informative?}
\label{sec:rq2}
\input{figures/success_breakdown}

\textbf{Behavioral metrics differentiate policy performance even when success rates are similar.}
When policies achieve comparable success, binary outcomes alone cannot distinguish their execution quality, yet behavioral metrics reveal meaningful differences.
In \texttt{Pack Box}, Pi-0.5 and ACT achieve overlapping success rates ($0.54 \pm 0.14$ vs.\ $0.48 \pm 0.14$), but ACT produces $4\times$ lower cartesian jerk ($6.8$ vs.\ $27.6$\,m/s$^3$) and $2.7\times$ shorter cartesian path lengths ($1.7$ vs.\ $4.6$\,m), as shown in Figure \ref{fig:packbox_diagnosis}(a) and (b).
Similarly, in \texttt{Lift Tray}, Pi-0.5 and ACT have overlapping success ($0.32 \pm 0.14$ vs.\ $0.26 \pm 0.14$), yet ACT achieves $5\times$ lower cartesian jerk ($6.1$ vs.\ $31.3$\,m/s$^3$) and $3.5\times$ shorter paths ($1.4$ vs.\ $4.8$\,m).
More broadly, Table~\ref{tab:metrics_summary_combined} shows that the highest-success model leads on only 33\% of the remaining metrics across all eight tasks.
Safety metrics such as slip count and environment collisions exhibit the most diversity, with three to four different models claiming the top rank across tasks.

\textbf{Outcome metrics diagnose policy behavior and expose skill gaps when success rates are low.}
When overall success is limited, outcome metrics pinpoint where failures accumulate.
In \texttt{Pack Box}, four of six policies achieve $\leq 12\%$ success, yet task progression separates them: Diffusion and Pi-0.5\,(LoRA) reach $0.27$ and $0.35$ respectively, indicating partial stage completion, while GR00T and xVLA stall at $0.07$ and $0.12$.
Stage-wise failure distributions (Figure~\ref{fig:packbox_diagnosis}(c)) reveal where each policy breaks down: GR00T and xVLA make no measurable progress in $66\%$ and $58\%$ of episodes respectively, whereas ACT's failures distribute more gradually across intermediate gripping and closing stages ($10$--$16\%$ per stage), indicating broader but shallower deficits rather than a single bottleneck.

\subsection{RQ3: How do the patterns and discriminative power of metrics vary across task domains and levels of complexity?}
\label{sec:rq3}

\input{figures/variation_difference}

\noindent \textbf{Outcome and safety metrics are the strongest discriminators between policies.}
In Figure~\ref{fig:variation_diff}, we report the coefficient of variation (CV) of each metric across policies, grouped by task variation (top) and base task (bottom). Higher CV indicates that a metric better separates policies.
Across both views, success rate is the single most discriminating metric (CV $\geq$ 1.0 in all conditions), followed by task progress (TP) and self-collision count (SCC). In contrast, trajectory length (TL) and slip count (SC) consistently exhibit low CV, suggesting they capture behaviors that are similar across policies regardless of task or variation.

\noindent \textbf{Metric discrimination is largely stable across task complexities.}
The variation-level analysis (Figure~\ref{fig:variation_diff}, top) shows no systematic decline in CV as variation complexity increases from static to combined position-and-orientation perturbations. This indicates that the metrics remain informative even under challenging conditions, rather than collapsing as all policies degrade uniformly.

\noindent \textbf{Discriminative power is task-dependent.}
The task-level analysis (Figure~\ref{fig:variation_diff}, bottom) reveals that the discriminative strength of individual metrics varies substantially across tasks. For example, SCC is especially discriminating on \texttt{RotateValve} and \texttt{StackTwoBlocks}, where collision-avoidance behavior diverges sharply between policies, while end-effector collision count (ECC) spikes on \texttt{LiftPot}. Efficiency metrics such as CPL and CT show moderate, relatively uniform discrimination across tasks, whereas coordination metrics (BAVD, BGVD) are most discriminating on \texttt{RotateValve} and \texttt{StackTwoBlocks}. This task-dependence reinforces that a comprehensive metric suite is necessary: any single metric would under-represent policy differences on certain tasks.

%% file: tables/task_performance_main.tex
\begin{table*}[htbp]
\centering
\footnotesize
\renewcommand{\arraystretch}{1.2}
\caption{\textbf{Summary of performance metrics across all methods on the combined variation.} Bold values indicate the best result for each metric within the task. Values are reported as mean $\pm$ 95\% confidence interval. The best performing method across each metric is highlighted in green. The final row reports human demonstration statistics (computed from successful teleoperated demos only), which provide a reference band for efficient execution.}
\begin{adjustbox}{width=\textwidth,center}
\begin{tabular}{lr|lr|lrlrlrlrlrlrlrlrlrlrlrlrlr}
\toprule
\textbf{Method} & \multicolumn{2}{|c|}{\textbf{Success $\uparrow$}} & \multicolumn{2}{c}{\textbf{TP $\uparrow$}} & \multicolumn{2}{c}{\textbf{BAVD $\downarrow$}} & \multicolumn{2}{c}{\textbf{BGVD $\downarrow$}} & \multicolumn{2}{c}{\textbf{CPL $\downarrow$}} & \multicolumn{2}{c}{\textbf{CT $\downarrow$}} & \multicolumn{2}{c}{\textbf{ECC $\downarrow$}} & \multicolumn{2}{c}{\textbf{JPL $\downarrow$}} & \multicolumn{2}{c}{\textbf{MCJ $\downarrow$}} & \multicolumn{2}{c}{\textbf{MJJ $\downarrow$}} & \multicolumn{2}{c}{\textbf{OPL $\downarrow$}} & \multicolumn{2}{c}{\textbf{SCC $\downarrow$}} & \multicolumn{2}{c}{\textbf{SC $\downarrow$}} & \multicolumn{2}{c}{\textbf{TL $\downarrow$}} \\
\midrule
\multicolumn{29}{c}{\textbf{Cube Handover}} \\
\midrule
\texttt{ACT} & \multicolumn{2}{|c|}{$0.08\,\pm\,0.11$} & \multicolumn{2}{c}{$0.23\,\pm\,0.08$} & \multicolumn{2}{c}{\cellcolor{ForestGreen!20}$\mathbf{0.18}\,\pm\,0.03$} & \multicolumn{2}{c}{$0.08\,\pm\,0.02$} & \multicolumn{2}{c}{\cellcolor{ForestGreen!20}$\mathbf{1.21}\,\pm\,0.12$} & \multicolumn{2}{c}{$5.70\,\pm\,0.42$} & \multicolumn{2}{c}{$1.72\,\pm\,0.31$} & \multicolumn{2}{c}{\cellcolor{ForestGreen!20}$\mathbf{8.70}\,\pm\,1.00$} & \multicolumn{2}{c}{\cellcolor{ForestGreen!20}$\mathbf{4.51}\,\pm\,0.80$} & \multicolumn{2}{c}{\cellcolor{ForestGreen!20}$\mathbf{23.60}\,\pm\,3.28$} & \multicolumn{2}{c}{\cellcolor{ForestGreen!20}$\mathbf{5.98}\,\pm\,0.80$} & \multicolumn{2}{c}{$6.28\,\pm\,1.55$} & \multicolumn{2}{c}{$0.28\,\pm\,0.15$} & \multicolumn{2}{c}{$941.02\,\pm\,56.95$} \\
\texttt{Diffusion} & \multicolumn{2}{|c|}{$0.06\,\pm\,0.10$} & \multicolumn{2}{c}{$0.27\,\pm\,0.07$} & \multicolumn{2}{c}{$0.32\,\pm\,0.03$} & \multicolumn{2}{c}{$0.13\,\pm\,0.01$} & \multicolumn{2}{c}{$5.07\,\pm\,0.35$} & \multicolumn{2}{c}{$4.80\,\pm\,0.28$} & \multicolumn{2}{c}{$2.94\,\pm\,0.49$} & \multicolumn{2}{c}{$26.14\,\pm\,2.02$} & \multicolumn{2}{c}{$24.73\,\pm\,0.90$} & \multicolumn{2}{c}{$133.00\,\pm\,4.88$} & \multicolumn{2}{c}{$19.61\,\pm\,1.59$} & \multicolumn{2}{c}{$2.46\,\pm\,0.57$} & \multicolumn{2}{c}{$0.22\,\pm\,0.14$} & \multicolumn{2}{c}{$939.76\,\pm\,52.04$} \\
\texttt{GR00T} & \multicolumn{2}{|c|}{$0.00\,\pm\,0.07$} & \multicolumn{2}{c}{$0.01\,\pm\,0.01$} & \multicolumn{2}{c}{$0.50\,\pm\,0.01$} & \multicolumn{2}{c}{$0.10\,\pm\,0.01$} & \multicolumn{2}{c}{$4.69\,\pm\,0.26$} & \multicolumn{2}{c}{$9.52\,\pm\,0.26$} & \multicolumn{2}{c}{$2.90\,\pm\,0.49$} & \multicolumn{2}{c}{$33.15\,\pm\,1.11$} & \multicolumn{2}{c}{$42.62\,\pm\,1.49$} & \multicolumn{2}{c}{$235.44\,\pm\,4.45$} & \multicolumn{2}{c}{$23.34\,\pm\,0.97$} & \multicolumn{2}{c}{$25.88\,\pm\,1.91$} & \multicolumn{2}{c}{\cellcolor{ForestGreen!20}$\mathbf{0.00}\,\pm\,0.07$} & \multicolumn{2}{c}{$976.10\,\pm\,23.48$} \\
\texttt{Pi\_0.5} & \multicolumn{2}{|c|}{\cellcolor{ForestGreen!20}$\mathbf{0.40}\,\pm\,0.14$} & \multicolumn{2}{c}{\cellcolor{ForestGreen!20}$\mathbf{0.65}\,\pm\,0.10$} & \multicolumn{2}{c}{$0.34\,\pm\,0.03$} & \multicolumn{2}{c}{\cellcolor{ForestGreen!20}$\mathbf{0.05}\,\pm\,0.01$} & \multicolumn{2}{c}{$2.22\,\pm\,0.44$} & \multicolumn{2}{c}{\cellcolor{ForestGreen!20}$\mathbf{3.66}\,\pm\,0.68$} & \multicolumn{2}{c}{\cellcolor{ForestGreen!20}$\mathbf{1.32}\,\pm\,0.47$} & \multicolumn{2}{c}{$11.22\,\pm\,2.31$} & \multicolumn{2}{c}{$18.14\,\pm\,1.78$} & \multicolumn{2}{c}{$88.53\,\pm\,7.98$} & \multicolumn{2}{c}{$8.00\,\pm\,1.68$} & \multicolumn{2}{c}{\cellcolor{ForestGreen!20}$\mathbf{2.30}\,\pm\,0.99$} & \multicolumn{2}{c}{$0.26\,\pm\,0.13$} & \multicolumn{2}{c}{\cellcolor{ForestGreen!20}$\mathbf{620.58}\,\pm\,120.85$} \\
\texttt{Pi\_0.5 (LoRA)} & \multicolumn{2}{|c|}{$0.10\,\pm\,0.11$} & \multicolumn{2}{c}{$0.29\,\pm\,0.08$} & \multicolumn{2}{c}{$0.40\,\pm\,0.03$} & \multicolumn{2}{c}{$0.08\,\pm\,0.01$} & \multicolumn{2}{c}{$4.61\,\pm\,0.49$} & \multicolumn{2}{c}{$5.22\,\pm\,0.42$} & \multicolumn{2}{c}{$4.22\,\pm\,0.63$} & \multicolumn{2}{c}{$26.98\,\pm\,3.04$} & \multicolumn{2}{c}{$27.40\,\pm\,1.99$} & \multicolumn{2}{c}{$153.46\,\pm\,11.11$} & \multicolumn{2}{c}{$18.32\,\pm\,2.03$} & \multicolumn{2}{c}{$5.44\,\pm\,2.24$} & \multicolumn{2}{c}{$0.20\,\pm\,0.15$} & \multicolumn{2}{c}{$900.28\,\pm\,71.03$} \\
\texttt{xVLA} & \multicolumn{2}{|c|}{$0.00\,\pm\,0.07$} & \multicolumn{2}{c}{$0.04\,\pm\,0.03$} & \multicolumn{2}{c}{$0.56\,\pm\,0.04$} & \multicolumn{2}{c}{$0.15\,\pm\,0.02$} & \multicolumn{2}{c}{$5.51\,\pm\,0.68$} & \multicolumn{2}{c}{$8.50\,\pm\,3.26$} & \multicolumn{2}{c}{$3.40\,\pm\,0.59$} & \multicolumn{2}{c}{$31.99\,\pm\,3.01$} & \multicolumn{2}{c}{$31.00\,\pm\,3.20$} & \multicolumn{2}{c}{$146.84\,\pm\,13.51$} & \multicolumn{2}{c}{$23.25\,\pm\,2.47$} & \multicolumn{2}{c}{$13.68\,\pm\,2.53$} & \multicolumn{2}{c}{$0.04\,\pm\,0.09$} & \multicolumn{2}{c}{$962.24\,\pm\,42.87$} \\
\midrule
\texttt{Human} & \multicolumn{2}{|c|}{$1.00\,\pm\,0.00$} & \multicolumn{2}{c}{$1.00\,\pm\,0.00$} & \multicolumn{2}{c}{$0.33\,\pm\,0.03$} & \multicolumn{2}{c}{$0.03\,\pm\,0.00$} & \multicolumn{2}{c}{$0.53\,\pm\,0.03$} & \multicolumn{2}{c}{$0.51\,\pm\,0.07$} & \multicolumn{2}{c}{$0.10\,\pm\,0.16$} & \multicolumn{2}{c}{$2.26\,\pm\,0.18$} & \multicolumn{2}{c}{$28.94\,\pm\,3.07$} & \multicolumn{2}{c}{$120.52\,\pm\,11.54$} & \multicolumn{2}{c}{$1.38\,\pm\,0.18$} & \multicolumn{2}{c}{$0.17\,\pm\,0.17$} & \multicolumn{2}{c}{$0.03\,\pm\,0.13$} & \multicolumn{2}{c}{$103.53\,\pm\,14.28$} \\
\midrule
\multicolumn{29}{c}{\textbf{Lift Pot}} \\
\midrule
\texttt{ACT} & \multicolumn{2}{|c|}{$0.44\,\pm\,0.14$} & \multicolumn{2}{c}{\cellcolor{ForestGreen!20}$\mathbf{0.58}\,\pm\,0.08$} & \multicolumn{2}{c}{\cellcolor{ForestGreen!20}$\mathbf{0.19}\,\pm\,0.02$} & \multicolumn{2}{c}{\cellcolor{ForestGreen!20}$\mathbf{0.07}\,\pm\,0.02$} & \multicolumn{2}{c}{\cellcolor{ForestGreen!20}$\mathbf{1.33}\,\pm\,0.27$} & \multicolumn{2}{c}{\cellcolor{ForestGreen!20}$\mathbf{3.32}\,\pm\,0.76$} & \multicolumn{2}{c}{\cellcolor{ForestGreen!20}$\mathbf{0.02}\,\pm\,0.08$} & \multicolumn{2}{c}{\cellcolor{ForestGreen!20}$\mathbf{7.82}\,\pm\,1.63$} & \multicolumn{2}{c}{\cellcolor{ForestGreen!20}$\mathbf{15.74}\,\pm\,2.57$} & \multicolumn{2}{c}{\cellcolor{ForestGreen!20}$\mathbf{82.66}\,\pm\,13.36$} & \multicolumn{2}{c}{\cellcolor{ForestGreen!20}$\mathbf{5.07}\,\pm\,0.98$} & \multicolumn{2}{c}{$0.98\,\pm\,0.78$} & \multicolumn{2}{c}{$0.28\,\pm\,0.18$} & \multicolumn{2}{c}{\cellcolor{ForestGreen!20}$\mathbf{518.30}\,\pm\,121.72$} \\
\texttt{Diffusion} & \multicolumn{2}{|c|}{$0.02\,\pm\,0.08$} & \multicolumn{2}{c}{$0.07\,\pm\,0.03$} & \multicolumn{2}{c}{$0.36\,\pm\,0.05$} & \multicolumn{2}{c}{$0.18\,\pm\,0.03$} & \multicolumn{2}{c}{$4.58\,\pm\,0.56$} & \multicolumn{2}{c}{$8.87\,\pm\,0.40$} & \multicolumn{2}{c}{$0.20\,\pm\,0.18$} & \multicolumn{2}{c}{$24.46\,\pm\,2.63$} & \multicolumn{2}{c}{$27.14\,\pm\,3.77$} & \multicolumn{2}{c}{$142.55\,\pm\,18.69$} & \multicolumn{2}{c}{$16.55\,\pm\,1.88$} & \multicolumn{2}{c}{$10.64\,\pm\,2.02$} & \multicolumn{2}{c}{\cellcolor{ForestGreen!20}$\mathbf{0.12}\,\pm\,0.12$} & \multicolumn{2}{c}{$978.06\,\pm\,32.64$} \\
\texttt{GR00T} & \multicolumn{2}{|c|}{$0.06\,\pm\,0.10$} & \multicolumn{2}{c}{$0.30\,\pm\,0.05$} & \multicolumn{2}{c}{$0.51\,\pm\,0.02$} & \multicolumn{2}{c}{$0.09\,\pm\,0.02$} & \multicolumn{2}{c}{$2.13\,\pm\,0.30$} & \multicolumn{2}{c}{$7.43\,\pm\,1.22$} & \multicolumn{2}{c}{$4.54\,\pm\,0.76$} & \multicolumn{2}{c}{$21.44\,\pm\,2.66$} & \multicolumn{2}{c}{$28.17\,\pm\,1.91$} & \multicolumn{2}{c}{$231.37\,\pm\,6.96$} & \multicolumn{2}{c}{$13.14\,\pm\,1.64$} & \multicolumn{2}{c}{$5.88\,\pm\,2.26$} & \multicolumn{2}{c}{$1.34\,\pm\,0.53$} & \multicolumn{2}{c}{$628.14\,\pm\,92.83$} \\
\texttt{Pi\_0.5} & \multicolumn{2}{|c|}{\cellcolor{ForestGreen!20}$\mathbf{0.52}\,\pm\,0.13$} & \multicolumn{2}{c}{$0.41\,\pm\,0.08$} & \multicolumn{2}{c}{$0.34\,\pm\,0.03$} & \multicolumn{2}{c}{$0.07\,\pm\,0.02$} & \multicolumn{2}{c}{$2.73\,\pm\,0.62$} & \multicolumn{2}{c}{$3.86\,\pm\,0.78$} & \multicolumn{2}{c}{$0.06\,\pm\,0.10$} & \multicolumn{2}{c}{$14.23\,\pm\,3.13$} & \multicolumn{2}{c}{$27.71\,\pm\,2.54$} & \multicolumn{2}{c}{$139.02\,\pm\,12.27$} & \multicolumn{2}{c}{$9.53\,\pm\,2.06$} & \multicolumn{2}{c}{\cellcolor{ForestGreen!20}$\mathbf{0.80}\,\pm\,0.67$} & \multicolumn{2}{c}{$0.28\,\pm\,0.15$} & \multicolumn{2}{c}{$550.78\,\pm\,117.31$} \\
\texttt{Pi\_0.5 (LoRA)} & \multicolumn{2}{|c|}{$0.08\,\pm\,0.11$} & \multicolumn{2}{c}{$0.22\,\pm\,0.07$} & \multicolumn{2}{c}{$0.61\,\pm\,0.03$} & \multicolumn{2}{c}{$0.20\,\pm\,0.03$} & \multicolumn{2}{c}{$8.65\,\pm\,0.91$} & \multicolumn{2}{c}{$6.17\,\pm\,0.59$} & \multicolumn{2}{c}{$0.58\,\pm\,0.32$} & \multicolumn{2}{c}{$47.03\,\pm\,4.83$} & \multicolumn{2}{c}{$53.22\,\pm\,2.58$} & \multicolumn{2}{c}{$294.12\,\pm\,12.29$} & \multicolumn{2}{c}{$28.68\,\pm\,2.85$} & \multicolumn{2}{c}{$5.56\,\pm\,1.82$} & \multicolumn{2}{c}{$0.28\,\pm\,0.14$} & \multicolumn{2}{c}{$854.18\,\pm\,80.23$} \\
\texttt{xVLA} & \multicolumn{2}{|c|}{$0.00\,\pm\,0.07$} & \multicolumn{2}{c}{$0.21\,\pm\,0.05$} & \multicolumn{2}{c}{$0.67\,\pm\,0.06$} & \multicolumn{2}{c}{$0.22\,\pm\,0.04$} & \multicolumn{2}{c}{$5.31\,\pm\,0.84$} & \multicolumn{2}{c}{$8.29\,\pm\,3.48$} & \multicolumn{2}{c}{$2.26\,\pm\,0.69$} & \multicolumn{2}{c}{$30.12\,\pm\,3.74$} & \multicolumn{2}{c}{$36.35\,\pm\,3.80$} & \multicolumn{2}{c}{$173.17\,\pm\,16.93$} & \multicolumn{2}{c}{$21.79\,\pm\,2.89$} & \multicolumn{2}{c}{$4.96\,\pm\,1.66$} & \multicolumn{2}{c}{$0.32\,\pm\,0.20$} & \multicolumn{2}{c}{$747.02\,\pm\,90.35$} \\
\midrule
\texttt{Human} & \multicolumn{2}{|c|}{$1.00\,\pm\,0.00$} & \multicolumn{2}{c}{$1.00\,\pm\,0.00$} & \multicolumn{2}{c}{$0.49\,\pm\,0.05$} & \multicolumn{2}{c}{$0.03\,\pm\,0.01$} & \multicolumn{2}{c}{$0.61\,\pm\,0.07$} & \multicolumn{2}{c}{$0.45\,\pm\,0.06$} & \multicolumn{2}{c}{$0.00\,\pm\,0.14$} & \multicolumn{2}{c}{$3.33\,\pm\,0.34$} & \multicolumn{2}{c}{$62.59\,\pm\,4.70$} & \multicolumn{2}{c}{$287.53\,\pm\,20.08$} & \multicolumn{2}{c}{$2.40\,\pm\,0.21$} & \multicolumn{2}{c}{$0.00\,\pm\,0.14$} & \multicolumn{2}{c}{$0.00\,\pm\,0.14$} & \multicolumn{2}{c}{$71.04\,\pm\,9.68$} \\
\midrule
\multicolumn{29}{c}{\textbf{Lift Tray}} \\
\midrule
\texttt{ACT} & \multicolumn{2}{|c|}{$0.26\,\pm\,0.14$} & \multicolumn{2}{c}{\cellcolor{ForestGreen!20}$\mathbf{0.81}\,\pm\,0.06$} & \multicolumn{2}{c}{\cellcolor{ForestGreen!20}$\mathbf{0.19}\,\pm\,0.02$} & \multicolumn{2}{c}{\cellcolor{ForestGreen!20}$\mathbf{0.06}\,\pm\,0.01$} & \multicolumn{2}{c}{\cellcolor{ForestGreen!20}$\mathbf{1.39}\,\pm\,0.19$} & \multicolumn{2}{c}{\cellcolor{ForestGreen!20}$\mathbf{5.63}\,\pm\,0.78$} & \multicolumn{2}{c}{\cellcolor{ForestGreen!20}$\mathbf{1.34}\,\pm\,0.57$} & \multicolumn{2}{c}{\cellcolor{ForestGreen!20}$\mathbf{10.43}\,\pm\,1.59$} & \multicolumn{2}{c}{\cellcolor{ForestGreen!20}$\mathbf{6.08}\,\pm\,0.64$} & \multicolumn{2}{c}{\cellcolor{ForestGreen!20}$\mathbf{40.51}\,\pm\,4.04$} & \multicolumn{2}{c}{\cellcolor{ForestGreen!20}$\mathbf{6.71}\,\pm\,1.06$} & \multicolumn{2}{c}{\cellcolor{ForestGreen!20}$\mathbf{2.26}\,\pm\,1.02$} & \multicolumn{2}{c}{$0.74\,\pm\,0.23$} & \multicolumn{2}{c}{$773.56\,\pm\,103.73$} \\
\texttt{Diffusion} & \multicolumn{2}{|c|}{$0.00\,\pm\,0.07$} & \multicolumn{2}{c}{$0.47\,\pm\,0.06$} & \multicolumn{2}{c}{$0.47\,\pm\,0.04$} & \multicolumn{2}{c}{$0.21\,\pm\,0.04$} & \multicolumn{2}{c}{$5.01\,\pm\,0.52$} & \multicolumn{2}{c}{$9.85\,\pm\,0.28$} & \multicolumn{2}{c}{$5.28\,\pm\,1.29$} & \multicolumn{2}{c}{$29.97\,\pm\,2.22$} & \multicolumn{2}{c}{$27.50\,\pm\,3.48$} & \multicolumn{2}{c}{$147.21\,\pm\,13.54$} & \multicolumn{2}{c}{$21.38\,\pm\,1.78$} & \multicolumn{2}{c}{$6.70\,\pm\,2.20$} & \multicolumn{2}{c}{$0.48\,\pm\,0.20$} & \multicolumn{2}{c}{$1000.00\,\pm\,0.00$} \\
\texttt{GR00T} & \multicolumn{2}{|c|}{$0.00\,\pm\,0.07$} & \multicolumn{2}{c}{$0.46\,\pm\,0.05$} & \multicolumn{2}{c}{$0.51\,\pm\,0.02$} & \multicolumn{2}{c}{$0.12\,\pm\,0.02$} & \multicolumn{2}{c}{$3.93\,\pm\,0.30$} & \multicolumn{2}{c}{$13.28\,\pm\,0.25$} & \multicolumn{2}{c}{$6.62\,\pm\,1.25$} & \multicolumn{2}{c}{$32.10\,\pm\,1.15$} & \multicolumn{2}{c}{$34.08\,\pm\,1.98$} & \multicolumn{2}{c}{$228.16\,\pm\,6.87$} & \multicolumn{2}{c}{$20.66\,\pm\,0.93$} & \multicolumn{2}{c}{$17.06\,\pm\,2.25$} & \multicolumn{2}{c}{$0.50\,\pm\,0.23$} & \multicolumn{2}{c}{$1000.00\,\pm\,0.00$} \\
\texttt{Pi\_0.5} & \multicolumn{2}{|c|}{\cellcolor{ForestGreen!20}$\mathbf{0.32}\,\pm\,0.14$} & \multicolumn{2}{c}{$0.75\,\pm\,0.07$} & \multicolumn{2}{c}{$0.52\,\pm\,0.05$} & \multicolumn{2}{c}{$0.14\,\pm\,0.03$} & \multicolumn{2}{c}{$4.83\,\pm\,0.84$} & \multicolumn{2}{c}{$5.87\,\pm\,0.94$} & \multicolumn{2}{c}{$2.62\,\pm\,0.99$} & \multicolumn{2}{c}{$29.67\,\pm\,5.06$} & \multicolumn{2}{c}{$31.27\,\pm\,2.31$} & \multicolumn{2}{c}{$181.95\,\pm\,11.42$} & \multicolumn{2}{c}{$21.82\,\pm\,3.78$} & \multicolumn{2}{c}{$5.22\,\pm\,2.07$} & \multicolumn{2}{c}{$0.92\,\pm\,0.22$} & \multicolumn{2}{c}{\cellcolor{ForestGreen!20}$\mathbf{719.16}\,\pm\,115.13$} \\
\texttt{Pi\_0.5 (LoRA)} & \multicolumn{2}{|c|}{$0.14\,\pm\,0.12$} & \multicolumn{2}{c}{$0.64\,\pm\,0.06$} & \multicolumn{2}{c}{$0.60\,\pm\,0.03$} & \multicolumn{2}{c}{$0.24\,\pm\,0.04$} & \multicolumn{2}{c}{$7.87\,\pm\,0.89$} & \multicolumn{2}{c}{$7.31\,\pm\,0.73$} & \multicolumn{2}{c}{$2.28\,\pm\,1.08$} & \multicolumn{2}{c}{$45.16\,\pm\,4.86$} & \multicolumn{2}{c}{$47.05\,\pm\,2.99$} & \multicolumn{2}{c}{$257.01\,\pm\,13.41$} & \multicolumn{2}{c}{$30.91\,\pm\,3.36$} & \multicolumn{2}{c}{$7.68\,\pm\,2.26$} & \multicolumn{2}{c}{$0.80\,\pm\,0.21$} & \multicolumn{2}{c}{$878.56\,\pm\,84.41$} \\
\texttt{xVLA} & \multicolumn{2}{|c|}{$0.00\,\pm\,0.07$} & \multicolumn{2}{c}{$0.35\,\pm\,0.04$} & \multicolumn{2}{c}{$0.63\,\pm\,0.06$} & \multicolumn{2}{c}{$0.24\,\pm\,0.04$} & \multicolumn{2}{c}{$7.01\,\pm\,0.92$} & \multicolumn{2}{c}{$10.49\,\pm\,0.43$} & \multicolumn{2}{c}{$4.88\,\pm\,0.94$} & \multicolumn{2}{c}{$37.46\,\pm\,3.72$} & \multicolumn{2}{c}{$32.82\,\pm\,4.39$} & \multicolumn{2}{c}{$151.68\,\pm\,16.81$} & \multicolumn{2}{c}{$28.46\,\pm\,3.09$} & \multicolumn{2}{c}{$10.46\,\pm\,2.68$} & \multicolumn{2}{c}{\cellcolor{ForestGreen!20}$\mathbf{0.16}\,\pm\,0.12$} & \multicolumn{2}{c}{$1000.00\,\pm\,0.00$} \\
\midrule
\texttt{Human} & \multicolumn{2}{|c|}{$1.00\,\pm\,0.00$} & \multicolumn{2}{c}{$1.00\,\pm\,0.00$} & \multicolumn{2}{c}{$0.39\,\pm\,0.04$} & \multicolumn{2}{c}{$0.03\,\pm\,0.00$} & \multicolumn{2}{c}{$0.52\,\pm\,0.03$} & \multicolumn{2}{c}{$0.56\,\pm\,0.03$} & \multicolumn{2}{c}{$0.00\,\pm\,0.06$} & \multicolumn{2}{c}{$2.87\,\pm\,0.13$} & \multicolumn{2}{c}{$26.77\,\pm\,2.93$} & \multicolumn{2}{c}{$153.00\,\pm\,16.82$} & \multicolumn{2}{c}{$2.04\,\pm\,0.08$} & \multicolumn{2}{c}{$0.00\,\pm\,0.06$} & \multicolumn{2}{c}{$0.00\,\pm\,0.06$} & \multicolumn{2}{c}{$83.62\,\pm\,3.96$} \\
\midrule
\multicolumn{29}{c}{\textbf{Pack Box}} \\
\midrule
\texttt{ACT} & \multicolumn{2}{|c|}{$0.48\,\pm\,0.13$} & \multicolumn{2}{c}{$0.46\,\pm\,0.08$} & \multicolumn{2}{c}{\cellcolor{ForestGreen!20}$\mathbf{0.25}\,\pm\,0.03$} & \multicolumn{2}{c}{\cellcolor{ForestGreen!20}$\mathbf{0.08}\,\pm\,0.02$} & \multicolumn{2}{c}{\cellcolor{ForestGreen!20}$\mathbf{1.68}\,\pm\,0.24$} & \multicolumn{2}{c}{\cellcolor{ForestGreen!20}$\mathbf{3.41}\,\pm\,0.62$} & \multicolumn{2}{c}{$0.74\,\pm\,0.37$} & \multicolumn{2}{c}{\cellcolor{ForestGreen!20}$\mathbf{7.87}\,\pm\,1.51$} & \multicolumn{2}{c}{\cellcolor{ForestGreen!20}$\mathbf{6.84}\,\pm\,1.09$} & \multicolumn{2}{c}{\cellcolor{ForestGreen!20}$\mathbf{29.15}\,\pm\,4.25$} & \multicolumn{2}{c}{\cellcolor{ForestGreen!20}$\mathbf{5.51}\,\pm\,1.16$} & \multicolumn{2}{c}{\cellcolor{ForestGreen!20}$\mathbf{0.72}\,\pm\,0.47$} & \multicolumn{2}{c}{$0.28\,\pm\,0.16$} & \multicolumn{2}{c}{$613.72\,\pm\,110.18$} \\
\texttt{Diffusion} & \multicolumn{2}{|c|}{$0.04\,\pm\,0.09$} & \multicolumn{2}{c}{$0.27\,\pm\,0.06$} & \multicolumn{2}{c}{$0.43\,\pm\,0.04$} & \multicolumn{2}{c}{$0.11\,\pm\,0.02$} & \multicolumn{2}{c}{$7.19\,\pm\,0.41$} & \multicolumn{2}{c}{$5.42\,\pm\,0.31$} & \multicolumn{2}{c}{\cellcolor{ForestGreen!20}$\mathbf{0.58}\,\pm\,0.34$} & \multicolumn{2}{c}{$36.50\,\pm\,2.51$} & \multicolumn{2}{c}{$29.98\,\pm\,2.07$} & \multicolumn{2}{c}{$154.68\,\pm\,11.16$} & \multicolumn{2}{c}{$25.85\,\pm\,1.81$} & \multicolumn{2}{c}{$9.82\,\pm\,3.05$} & \multicolumn{2}{c}{$0.44\,\pm\,0.16$} & \multicolumn{2}{c}{$962.98\,\pm\,44.27$} \\
\texttt{GR00T} & \multicolumn{2}{|c|}{$0.00\,\pm\,0.07$} & \multicolumn{2}{c}{$0.07\,\pm\,0.03$} & \multicolumn{2}{c}{$0.50\,\pm\,0.02$} & \multicolumn{2}{c}{$0.24\,\pm\,0.04$} & \multicolumn{2}{c}{$4.77\,\pm\,0.36$} & \multicolumn{2}{c}{$10.75\,\pm\,0.20$} & \multicolumn{2}{c}{$1.04\,\pm\,0.47$} & \multicolumn{2}{c}{$29.29\,\pm\,1.37$} & \multicolumn{2}{c}{$40.49\,\pm\,3.08$} & \multicolumn{2}{c}{$207.72\,\pm\,9.51$} & \multicolumn{2}{c}{$19.50\,\pm\,1.10$} & \multicolumn{2}{c}{$13.30\,\pm\,2.92$} & \multicolumn{2}{c}{$0.30\,\pm\,0.19$} & \multicolumn{2}{c}{$1000.00\,\pm\,0.00$} \\
\texttt{Pi\_0.5} & \multicolumn{2}{|c|}{\cellcolor{ForestGreen!20}$\mathbf{0.54}\,\pm\,0.14$} & \multicolumn{2}{c}{\cellcolor{ForestGreen!20}$\mathbf{0.55}\,\pm\,0.06$} & \multicolumn{2}{c}{$0.73\,\pm\,0.07$} & \multicolumn{2}{c}{$0.10\,\pm\,0.01$} & \multicolumn{2}{c}{$4.60\,\pm\,0.89$} & \multicolumn{2}{c}{$3.70\,\pm\,0.72$} & \multicolumn{2}{c}{$0.90\,\pm\,0.41$} & \multicolumn{2}{c}{$22.61\,\pm\,4.56$} & \multicolumn{2}{c}{$27.59\,\pm\,1.39$} & \multicolumn{2}{c}{$130.35\,\pm\,7.35$} & \multicolumn{2}{c}{$15.84\,\pm\,3.25$} & \multicolumn{2}{c}{$1.24\,\pm\,1.19$} & \multicolumn{2}{c}{$0.44\,\pm\,0.20$} & \multicolumn{2}{c}{\cellcolor{ForestGreen!20}$\mathbf{566.54}\,\pm\,113.98$} \\
\texttt{Pi\_0.5 (LoRA)} & \multicolumn{2}{|c|}{$0.12\,\pm\,0.12$} & \multicolumn{2}{c}{$0.35\,\pm\,0.06$} & \multicolumn{2}{c}{$0.68\,\pm\,0.04$} & \multicolumn{2}{c}{$0.17\,\pm\,0.03$} & \multicolumn{2}{c}{$7.92\,\pm\,0.65$} & \multicolumn{2}{c}{$6.01\,\pm\,0.44$} & \multicolumn{2}{c}{$1.18\,\pm\,0.47$} & \multicolumn{2}{c}{$45.24\,\pm\,3.63$} & \multicolumn{2}{c}{$42.69\,\pm\,2.35$} & \multicolumn{2}{c}{$225.24\,\pm\,11.40$} & \multicolumn{2}{c}{$31.91\,\pm\,2.61$} & \multicolumn{2}{c}{$7.54\,\pm\,2.64$} & \multicolumn{2}{c}{$0.62\,\pm\,0.23$} & \multicolumn{2}{c}{$923.82\,\pm\,62.68$} \\
\texttt{xVLA} & \multicolumn{2}{|c|}{$0.02\,\pm\,0.08$} & \multicolumn{2}{c}{$0.12\,\pm\,0.05$} & \multicolumn{2}{c}{$0.79\,\pm\,0.05$} & \multicolumn{2}{c}{$0.25\,\pm\,0.03$} & \multicolumn{2}{c}{$9.76\,\pm\,0.88$} & \multicolumn{2}{c}{$7.29\,\pm\,0.35$} & \multicolumn{2}{c}{$1.30\,\pm\,0.53$} & \multicolumn{2}{c}{$47.57\,\pm\,3.10$} & \multicolumn{2}{c}{$46.60\,\pm\,3.63$} & \multicolumn{2}{c}{$200.34\,\pm\,12.90$} & \multicolumn{2}{c}{$34.53\,\pm\,2.36$} & \multicolumn{2}{c}{$9.24\,\pm\,3.08$} & \multicolumn{2}{c}{\cellcolor{ForestGreen!20}$\mathbf{0.14}\,\pm\,0.11$} & \multicolumn{2}{c}{$988.12\,\pm\,23.28$} \\
\midrule
\texttt{Human} & \multicolumn{2}{|c|}{$1.00\,\pm\,0.00$} & \multicolumn{2}{c}{$1.00\,\pm\,0.00$} & \multicolumn{2}{c}{$0.51\,\pm\,0.04$} & \multicolumn{2}{c}{$0.05\,\pm\,0.00$} & \multicolumn{2}{c}{$1.30\,\pm\,0.06$} & \multicolumn{2}{c}{$0.68\,\pm\,0.04$} & \multicolumn{2}{c}{$0.84\,\pm\,0.37$} & \multicolumn{2}{c}{$5.38\,\pm\,0.34$} & \multicolumn{2}{c}{$28.24\,\pm\,1.50$} & \multicolumn{2}{c}{$134.77\,\pm\,6.73$} & \multicolumn{2}{c}{$3.35\,\pm\,0.21$} & \multicolumn{2}{c}{$0.18\,\pm\,0.16$} & \multicolumn{2}{c}{$0.03\,\pm\,0.07$} & \multicolumn{2}{c}{$129.66\,\pm\,8.00$} \\
\midrule
\multicolumn{29}{c}{\textbf{Pick Single Book From Table}} \\
\midrule
\texttt{ACT} & \multicolumn{2}{|c|}{\cellcolor{ForestGreen!20}$\mathbf{0.34}\,\pm\,0.14$} & \multicolumn{2}{c}{\cellcolor{ForestGreen!20}$\mathbf{0.55}\,\pm\,0.13$} & \multicolumn{2}{c}{\cellcolor{ForestGreen!20}$\mathbf{0.51}\,\pm\,0.09$} & \multicolumn{2}{c}{$0.25\,\pm\,0.02$} & \multicolumn{2}{c}{\cellcolor{ForestGreen!20}$\mathbf{1.13}\,\pm\,0.24$} & \multicolumn{2}{c}{\cellcolor{ForestGreen!20}$\mathbf{5.79}\,\pm\,1.41$} & \multicolumn{2}{c}{\cellcolor{ForestGreen!20}$\mathbf{1.58}\,\pm\,0.46$} & \multicolumn{2}{c}{\cellcolor{ForestGreen!20}$\mathbf{6.70}\,\pm\,1.43$} & \multicolumn{2}{c}{\cellcolor{ForestGreen!20}$\mathbf{8.59}\,\pm\,1.32$} & \multicolumn{2}{c}{\cellcolor{ForestGreen!20}$\mathbf{39.56}\,\pm\,5.31$} & \multicolumn{2}{c}{\cellcolor{ForestGreen!20}$\mathbf{3.33}\,\pm\,0.70$} & \multicolumn{2}{c}{\cellcolor{ForestGreen!20}$\mathbf{0.18}\,\pm\,0.14$} & \multicolumn{2}{c}{$0.10\,\pm\,0.11$} & \multicolumn{2}{c}{$504.46\,\pm\,121.23$} \\
\texttt{Diffusion} & \multicolumn{2}{|c|}{$0.02\,\pm\,0.08$} & \multicolumn{2}{c}{$0.15\,\pm\,0.09$} & \multicolumn{2}{c}{$0.59\,\pm\,0.04$} & \multicolumn{2}{c}{$0.24\,\pm\,0.03$} & \multicolumn{2}{c}{$4.42\,\pm\,0.49$} & \multicolumn{2}{c}{$11.20\,\pm\,0.83$} & \multicolumn{2}{c}{$5.52\,\pm\,1.27$} & \multicolumn{2}{c}{$28.17\,\pm\,2.59$} & \multicolumn{2}{c}{$21.24\,\pm\,1.98$} & \multicolumn{2}{c}{$136.43\,\pm\,10.11$} & \multicolumn{2}{c}{$17.38\,\pm\,1.69$} & \multicolumn{2}{c}{$4.26\,\pm\,1.50$} & \multicolumn{2}{c}{$0.04\,\pm\,0.09$} & \multicolumn{2}{c}{$926.04\,\pm\,61.77$} \\
\texttt{GR00T} & \multicolumn{2}{|c|}{$0.00\,\pm\,0.07$} & \multicolumn{2}{c}{$0.05\,\pm\,0.05$} & \multicolumn{2}{c}{$0.56\,\pm\,0.03$} & \multicolumn{2}{c}{$0.26\,\pm\,0.03$} & \multicolumn{2}{c}{$2.96\,\pm\,0.36$} & \multicolumn{2}{c}{$15.47\,\pm\,1.63$} & \multicolumn{2}{c}{$9.12\,\pm\,1.30$} & \multicolumn{2}{c}{$28.12\,\pm\,2.69$} & \multicolumn{2}{c}{$28.97\,\pm\,1.90$} & \multicolumn{2}{c}{$237.89\,\pm\,7.47$} & \multicolumn{2}{c}{$18.37\,\pm\,1.90$} & \multicolumn{2}{c}{$10.90\,\pm\,2.44$} & \multicolumn{2}{c}{$0.04\,\pm\,0.09$} & \multicolumn{2}{c}{$821.42\,\pm\,83.40$} \\
\texttt{Pi\_0.5} & \multicolumn{2}{|c|}{$0.22\,\pm\,0.13$} & \multicolumn{2}{c}{$0.44\,\pm\,0.13$} & \multicolumn{2}{c}{$0.72\,\pm\,0.07$} & \multicolumn{2}{c}{\cellcolor{ForestGreen!20}$\mathbf{0.20}\,\pm\,0.02$} & \multicolumn{2}{c}{$1.89\,\pm\,0.47$} & \multicolumn{2}{c}{$6.03\,\pm\,1.54$} & \multicolumn{2}{c}{$3.02\,\pm\,0.88$} & \multicolumn{2}{c}{$11.37\,\pm\,2.88$} & \multicolumn{2}{c}{$17.94\,\pm\,1.51$} & \multicolumn{2}{c}{$91.87\,\pm\,6.23$} & \multicolumn{2}{c}{$6.67\,\pm\,1.76$} & \multicolumn{2}{c}{$0.28\,\pm\,0.24$} & \multicolumn{2}{c}{$0.12\,\pm\,0.11$} & \multicolumn{2}{c}{\cellcolor{ForestGreen!20}$\mathbf{458.76}\,\pm\,120.93$} \\
\texttt{Pi\_0.5 (LoRA)} & \multicolumn{2}{|c|}{$0.00\,\pm\,0.07$} & \multicolumn{2}{c}{$0.22\,\pm\,0.09$} & \multicolumn{2}{c}{$0.67\,\pm\,0.08$} & \multicolumn{2}{c}{$0.20\,\pm\,0.02$} & \multicolumn{2}{c}{$3.65\,\pm\,0.64$} & \multicolumn{2}{c}{$9.11\,\pm\,1.46$} & \multicolumn{2}{c}{$5.80\,\pm\,1.44$} & \multicolumn{2}{c}{$24.05\,\pm\,4.15$} & \multicolumn{2}{c}{$26.09\,\pm\,1.80$} & \multicolumn{2}{c}{$156.80\,\pm\,9.46$} & \multicolumn{2}{c}{$14.76\,\pm\,2.55$} & \multicolumn{2}{c}{$2.36\,\pm\,1.38$} & \multicolumn{2}{c}{$0.10\,\pm\,0.11$} & \multicolumn{2}{c}{$705.64\,\pm\,114.01$} \\
\texttt{xVLA} & \multicolumn{2}{|c|}{$0.00\,\pm\,0.07$} & \multicolumn{2}{c}{$0.02\,\pm\,0.08$} & \multicolumn{2}{c}{$0.72\,\pm\,0.06$} & \multicolumn{2}{c}{$0.21\,\pm\,0.03$} & \multicolumn{2}{c}{$5.10\,\pm\,0.72$} & \multicolumn{2}{c}{$13.53\,\pm\,0.81$} & \multicolumn{2}{c}{$5.58\,\pm\,0.96$} & \multicolumn{2}{c}{$31.56\,\pm\,2.94$} & \multicolumn{2}{c}{$22.35\,\pm\,2.75$} & \multicolumn{2}{c}{$114.50\,\pm\,11.31$} & \multicolumn{2}{c}{$22.67\,\pm\,2.35$} & \multicolumn{2}{c}{$7.90\,\pm\,1.75$} & \multicolumn{2}{c}{\cellcolor{ForestGreen!20}$\mathbf{0.02}\,\pm\,0.08$} & \multicolumn{2}{c}{$949.08\,\pm\,47.67$} \\
\midrule
\texttt{Human} & \multicolumn{2}{|c|}{$1.00\,\pm\,0.00$} & \multicolumn{2}{c}{$1.00\,\pm\,0.00$} & \multicolumn{2}{c}{$0.87\,\pm\,0.00$} & \multicolumn{2}{c}{$0.18\,\pm\,0.00$} & \multicolumn{2}{c}{$0.76\,\pm\,0.00$} & \multicolumn{2}{c}{$1.20\,\pm\,0.00$} & \multicolumn{2}{c}{$0.00\,\pm\,0.00$} & \multicolumn{2}{c}{$4.09\,\pm\,0.00$} & \multicolumn{2}{c}{$30.32\,\pm\,0.00$} & \multicolumn{2}{c}{$138.21\,\pm\,0.00$} & \multicolumn{2}{c}{$1.90\,\pm\,0.00$} & \multicolumn{2}{c}{$0.00\,\pm\,0.00$} & \multicolumn{2}{c}{$0.00\,\pm\,0.00$} & \multicolumn{2}{c}{$118.00\,\pm\,0.00$} \\
\midrule
\multicolumn{29}{c}{\textbf{Rotate Valve}} \\
\midrule
\texttt{ACT} & \multicolumn{2}{|c|}{$0.36\,\pm\,0.14$} & \multicolumn{2}{c}{$0.57\,\pm\,0.10$} & \multicolumn{2}{c}{\cellcolor{ForestGreen!20}$\mathbf{0.17}\,\pm\,0.02$} & \multicolumn{2}{c}{$0.04\,\pm\,0.01$} & \multicolumn{2}{c}{\cellcolor{ForestGreen!20}$\mathbf{0.40}\,\pm\,0.06$} & \multicolumn{2}{c}{$11.09\,\pm\,1.15$} & \multicolumn{2}{c}{$10.46\,\pm\,1.16$} & \multicolumn{2}{c}{\cellcolor{ForestGreen!20}$\mathbf{2.91}\,\pm\,0.45$} & \multicolumn{2}{c}{\cellcolor{ForestGreen!20}$\mathbf{0.99}\,\pm\,0.12$} & \multicolumn{2}{c}{\cellcolor{ForestGreen!20}$\mathbf{5.39}\,\pm\,0.62$} & \multicolumn{2}{c}{\cellcolor{ForestGreen!20}$\mathbf{2.23}\,\pm\,0.35$} & \multicolumn{2}{c}{$0.44\,\pm\,0.38$} & \multicolumn{2}{c}{\cellcolor{ForestGreen!20}$\mathbf{0.00}\,\pm\,0.07$} & \multicolumn{2}{c}{$828.94\,\pm\,78.48$} \\
\texttt{Diffusion} & \multicolumn{2}{|c|}{$0.04\,\pm\,0.09$} & \multicolumn{2}{c}{$0.23\,\pm\,0.07$} & \multicolumn{2}{c}{$0.44\,\pm\,0.04$} & \multicolumn{2}{c}{$0.10\,\pm\,0.02$} & \multicolumn{2}{c}{$2.10\,\pm\,0.23$} & \multicolumn{2}{c}{$11.62\,\pm\,0.52$} & \multicolumn{2}{c}{\cellcolor{ForestGreen!20}$\mathbf{10.18}\,\pm\,1.47$} & \multicolumn{2}{c}{$13.61\,\pm\,1.18$} & \multicolumn{2}{c}{$5.88\,\pm\,0.54$} & \multicolumn{2}{c}{$41.95\,\pm\,3.08$} & \multicolumn{2}{c}{$9.76\,\pm\,0.94$} & \multicolumn{2}{c}{$2.86\,\pm\,1.06$} & \multicolumn{2}{c}{\cellcolor{ForestGreen!20}$\mathbf{0.00}\,\pm\,0.07$} & \multicolumn{2}{c}{$977.16\,\pm\,35.82$} \\
\texttt{GR00T} & \multicolumn{2}{|c|}{$0.00\,\pm\,0.07$} & \multicolumn{2}{c}{$0.11\,\pm\,0.04$} & \multicolumn{2}{c}{$0.63\,\pm\,0.03$} & \multicolumn{2}{c}{$0.08\,\pm\,0.01$} & \multicolumn{2}{c}{$2.69\,\pm\,0.17$} & \multicolumn{2}{c}{$18.11\,\pm\,0.63$} & \multicolumn{2}{c}{$23.70\,\pm\,2.64$} & \multicolumn{2}{c}{$21.46\,\pm\,0.77$} & \multicolumn{2}{c}{$14.40\,\pm\,0.81$} & \multicolumn{2}{c}{$88.73\,\pm\,2.50$} & \multicolumn{2}{c}{$15.09\,\pm\,0.75$} & \multicolumn{2}{c}{$16.40\,\pm\,2.50$} & \multicolumn{2}{c}{\cellcolor{ForestGreen!20}$\mathbf{0.00}\,\pm\,0.07$} & \multicolumn{2}{c}{$1000.00\,\pm\,0.00$} \\
\texttt{Pi\_0.5} & \multicolumn{2}{|c|}{\cellcolor{ForestGreen!20}$\mathbf{0.72}\,\pm\,0.14$} & \multicolumn{2}{c}{\cellcolor{ForestGreen!20}$\mathbf{0.88}\,\pm\,0.05$} & \multicolumn{2}{c}{$0.36\,\pm\,0.02$} & \multicolumn{2}{c}{\cellcolor{ForestGreen!20}$\mathbf{0.03}\,\pm\,0.00$} & \multicolumn{2}{c}{$0.62\,\pm\,0.14$} & \multicolumn{2}{c}{\cellcolor{ForestGreen!20}$\mathbf{8.65}\,\pm\,1.59$} & \multicolumn{2}{c}{$12.42\,\pm\,2.36$} & \multicolumn{2}{c}{$4.67\,\pm\,0.95$} & \multicolumn{2}{c}{$2.99\,\pm\,0.22$} & \multicolumn{2}{c}{$19.81\,\pm\,1.29$} & \multicolumn{2}{c}{$3.43\,\pm\,0.69$} & \multicolumn{2}{c}{\cellcolor{ForestGreen!20}$\mathbf{0.02}\,\pm\,0.08$} & \multicolumn{2}{c}{\cellcolor{ForestGreen!20}$\mathbf{0.00}\,\pm\,0.07$} & \multicolumn{2}{c}{\cellcolor{ForestGreen!20}$\mathbf{527.64}\,\pm\,95.25$} \\
\texttt{Pi\_0.5 (LoRA)} & \multicolumn{2}{|c|}{$0.28\,\pm\,0.14$} & \multicolumn{2}{c}{$0.62\,\pm\,0.08$} & \multicolumn{2}{c}{$0.32\,\pm\,0.03$} & \multicolumn{2}{c}{$0.06\,\pm\,0.01$} & \multicolumn{2}{c}{$1.19\,\pm\,0.16$} & \multicolumn{2}{c}{$13.09\,\pm\,1.21$} & \multicolumn{2}{c}{$19.80\,\pm\,2.58$} & \multicolumn{2}{c}{$8.85\,\pm\,1.15$} & \multicolumn{2}{c}{$5.22\,\pm\,0.44$} & \multicolumn{2}{c}{$34.89\,\pm\,2.58$} & \multicolumn{2}{c}{$5.70\,\pm\,0.72$} & \multicolumn{2}{c}{$0.16\,\pm\,0.20$} & \multicolumn{2}{c}{\cellcolor{ForestGreen!20}$\mathbf{0.00}\,\pm\,0.07$} & \multicolumn{2}{c}{$853.50\,\pm\,76.04$} \\
\texttt{xVLA} & \multicolumn{2}{|c|}{$0.10\,\pm\,0.11$} & \multicolumn{2}{c}{$0.15\,\pm\,0.06$} & \multicolumn{2}{c}{$0.73\,\pm\,0.05$} & \multicolumn{2}{c}{$0.13\,\pm\,0.02$} & \multicolumn{2}{c}{$3.07\,\pm\,0.29$} & \multicolumn{2}{c}{$11.74\,\pm\,0.62$} & \multicolumn{2}{c}{$12.34\,\pm\,2.19$} & \multicolumn{2}{c}{$19.40\,\pm\,1.35$} & \multicolumn{2}{c}{$7.30\,\pm\,0.55$} & \multicolumn{2}{c}{$38.85\,\pm\,2.53$} & \multicolumn{2}{c}{$14.54\,\pm\,1.06$} & \multicolumn{2}{c}{$2.40\,\pm\,0.91$} & \multicolumn{2}{c}{\cellcolor{ForestGreen!20}$\mathbf{0.00}\,\pm\,0.07$} & \multicolumn{2}{c}{$953.00\,\pm\,44.60$} \\
\midrule
\texttt{Human} & \multicolumn{2}{|c|}{$1.00\,\pm\,0.00$} & \multicolumn{2}{c}{$1.00\,\pm\,0.00$} & \multicolumn{2}{c}{$0.37\,\pm\,0.04$} & \multicolumn{2}{c}{$0.01\,\pm\,0.00$} & \multicolumn{2}{c}{$0.20\,\pm\,0.02$} & \multicolumn{2}{c}{$1.98\,\pm\,0.22$} & \multicolumn{2}{c}{$6.38\,\pm\,0.73$} & \multicolumn{2}{c}{$1.24\,\pm\,0.11$} & \multicolumn{2}{c}{$3.60\,\pm\,0.31$} & \multicolumn{2}{c}{$22.26\,\pm\,2.06$} & \multicolumn{2}{c}{$1.04\,\pm\,0.10$} & \multicolumn{2}{c}{$0.00\,\pm\,0.19$} & \multicolumn{2}{c}{$0.00\,\pm\,0.19$} & \multicolumn{2}{c}{$150.12\,\pm\,16.61$} \\
\midrule
\multicolumn{29}{c}{\textbf{Stack Single Book Shelf}} \\
\midrule
\texttt{ACT} & \multicolumn{2}{|c|}{$0.08\,\pm\,0.11$} & \multicolumn{2}{c}{$0.15\,\pm\,0.08$} & \multicolumn{2}{c}{\cellcolor{ForestGreen!20}$\mathbf{0.50}\,\pm\,0.07$} & \multicolumn{2}{c}{$0.22\,\pm\,0.02$} & \multicolumn{2}{c}{\cellcolor{ForestGreen!20}$\mathbf{1.26}\,\pm\,0.28$} & \multicolumn{2}{c}{$6.56\,\pm\,1.40$} & \multicolumn{2}{c}{$3.66\,\pm\,0.86$} & \multicolumn{2}{c}{\cellcolor{ForestGreen!20}$\mathbf{8.91}\,\pm\,1.99$} & \multicolumn{2}{c}{\cellcolor{ForestGreen!20}$\mathbf{8.61}\,\pm\,1.38$} & \multicolumn{2}{c}{\cellcolor{ForestGreen!20}$\mathbf{39.15}\,\pm\,5.69$} & \multicolumn{2}{c}{\cellcolor{ForestGreen!20}$\mathbf{5.33}\,\pm\,1.12$} & \multicolumn{2}{c}{\cellcolor{ForestGreen!20}$\mathbf{0.80}\,\pm\,0.61$} & \multicolumn{2}{c}{$0.08\,\pm\,0.11$} & \multicolumn{2}{c}{$560.22\,\pm\,120.10$} \\
\texttt{Diffusion} & \multicolumn{2}{|c|}{$0.00\,\pm\,0.07$} & \multicolumn{2}{c}{$0.01\,\pm\,0.01$} & \multicolumn{2}{c}{$0.78\,\pm\,0.06$} & \multicolumn{2}{c}{$0.24\,\pm\,0.02$} & \multicolumn{2}{c}{$6.01\,\pm\,0.63$} & \multicolumn{2}{c}{$11.95\,\pm\,0.62$} & \multicolumn{2}{c}{$5.74\,\pm\,0.83$} & \multicolumn{2}{c}{$35.11\,\pm\,2.58$} & \multicolumn{2}{c}{$29.84\,\pm\,3.97$} & \multicolumn{2}{c}{$166.93\,\pm\,15.36$} & \multicolumn{2}{c}{$22.73\,\pm\,1.84$} & \multicolumn{2}{c}{$3.54\,\pm\,1.22$} & \multicolumn{2}{c}{\cellcolor{ForestGreen!20}$\mathbf{0.00}\,\pm\,0.07$} & \multicolumn{2}{c}{$966.84\,\pm\,45.49$} \\
\texttt{GR00T} & \multicolumn{2}{|c|}{$0.00\,\pm\,0.07$} & \multicolumn{2}{c}{$0.04\,\pm\,0.03$} & \multicolumn{2}{c}{$0.56\,\pm\,0.02$} & \multicolumn{2}{c}{$0.29\,\pm\,0.03$} & \multicolumn{2}{c}{$3.17\,\pm\,0.41$} & \multicolumn{2}{c}{$16.17\,\pm\,1.54$} & \multicolumn{2}{c}{$9.46\,\pm\,1.34$} & \multicolumn{2}{c}{$28.67\,\pm\,2.49$} & \multicolumn{2}{c}{$31.38\,\pm\,2.57$} & \multicolumn{2}{c}{$236.75\,\pm\,5.66$} & \multicolumn{2}{c}{$18.50\,\pm\,1.71$} & \multicolumn{2}{c}{$11.38\,\pm\,2.70$} & \multicolumn{2}{c}{$0.04\,\pm\,0.09$} & \multicolumn{2}{c}{$842.26\,\pm\,76.69$} \\
\texttt{Pi\_0.5} & \multicolumn{2}{|c|}{\cellcolor{ForestGreen!20}$\mathbf{0.12}\,\pm\,0.12$} & \multicolumn{2}{c}{\cellcolor{ForestGreen!20}$\mathbf{0.27}\,\pm\,0.10$} & \multicolumn{2}{c}{$0.79\,\pm\,0.06$} & \multicolumn{2}{c}{\cellcolor{ForestGreen!20}$\mathbf{0.20}\,\pm\,0.01$} & \multicolumn{2}{c}{$2.04\,\pm\,0.56$} & \multicolumn{2}{c}{\cellcolor{ForestGreen!20}$\mathbf{5.96}\,\pm\,1.60$} & \multicolumn{2}{c}{\cellcolor{ForestGreen!20}$\mathbf{2.16}\,\pm\,0.58$} & \multicolumn{2}{c}{$12.82\,\pm\,3.59$} & \multicolumn{2}{c}{$19.90\,\pm\,1.25$} & \multicolumn{2}{c}{$104.69\,\pm\,5.90$} & \multicolumn{2}{c}{$7.39\,\pm\,2.09$} & \multicolumn{2}{c}{$2.16\,\pm\,1.45$} & \multicolumn{2}{c}{$0.16\,\pm\,0.17$} & \multicolumn{2}{c}{\cellcolor{ForestGreen!20}$\mathbf{414.04}\,\pm\,114.28$} \\
\texttt{Pi\_0.5 (LoRA)} & \multicolumn{2}{|c|}{$0.00\,\pm\,0.07$} & \multicolumn{2}{c}{$0.05\,\pm\,0.03$} & \multicolumn{2}{c}{$0.65\,\pm\,0.06$} & \multicolumn{2}{c}{$0.21\,\pm\,0.02$} & \multicolumn{2}{c}{$3.60\,\pm\,0.72$} & \multicolumn{2}{c}{$9.15\,\pm\,1.67$} & \multicolumn{2}{c}{$5.76\,\pm\,1.28$} & \multicolumn{2}{c}{$23.72\,\pm\,4.75$} & \multicolumn{2}{c}{$27.81\,\pm\,2.03$} & \multicolumn{2}{c}{$163.44\,\pm\,11.25$} & \multicolumn{2}{c}{$14.69\,\pm\,2.95$} & \multicolumn{2}{c}{$1.36\,\pm\,0.68$} & \multicolumn{2}{c}{$0.04\,\pm\,0.09$} & \multicolumn{2}{c}{$653.92\,\pm\,120.31$} \\
\texttt{xVLA} & \multicolumn{2}{|c|}{$0.00\,\pm\,0.07$} & \multicolumn{2}{c}{$0.01\,\pm\,0.01$} & \multicolumn{2}{c}{$0.78\,\pm\,0.06$} & \multicolumn{2}{c}{$0.24\,\pm\,0.04$} & \multicolumn{2}{c}{$5.53\,\pm\,0.77$} & \multicolumn{2}{c}{$13.64\,\pm\,0.82$} & \multicolumn{2}{c}{$7.28\,\pm\,1.31$} & \multicolumn{2}{c}{$33.77\,\pm\,3.27$} & \multicolumn{2}{c}{$29.08\,\pm\,3.57$} & \multicolumn{2}{c}{$150.93\,\pm\,15.57$} & \multicolumn{2}{c}{$24.37\,\pm\,2.47$} & \multicolumn{2}{c}{$10.16\,\pm\,2.45$} & \multicolumn{2}{c}{$0.02\,\pm\,0.08$} & \multicolumn{2}{c}{$937.06\,\pm\,49.54$} \\
\midrule
\texttt{Human} & \multicolumn{2}{|c|}{$1.00\,\pm\,0.00$} & \multicolumn{2}{c}{$1.00\,\pm\,0.00$} & \multicolumn{2}{c}{$0.85\,\pm\,0.00$} & \multicolumn{2}{c}{$0.20\,\pm\,0.00$} & \multicolumn{2}{c}{$1.10\,\pm\,0.00$} & \multicolumn{2}{c}{$2.12\,\pm\,0.00$} & \multicolumn{2}{c}{$0.00\,\pm\,0.00$} & \multicolumn{2}{c}{$6.29\,\pm\,0.00$} & \multicolumn{2}{c}{$19.79\,\pm\,0.00$} & \multicolumn{2}{c}{$100.20\,\pm\,0.00$} & \multicolumn{2}{c}{$2.90\,\pm\,0.00$} & \multicolumn{2}{c}{$0.00\,\pm\,0.00$} & \multicolumn{2}{c}{$0.00\,\pm\,0.00$} & \multicolumn{2}{c}{$194.00\,\pm\,0.00$} \\
\midrule
\multicolumn{29}{c}{\textbf{Stack Two Blocks}} \\
\midrule
\texttt{ACT} & \multicolumn{2}{|c|}{$0.18\,\pm\,0.13$} & \multicolumn{2}{c}{$0.28\,\pm\,0.07$} & \multicolumn{2}{c}{\cellcolor{ForestGreen!20}$\mathbf{0.06}\,\pm\,0.01$} & \multicolumn{2}{c}{$0.08\,\pm\,0.01$} & \multicolumn{2}{c}{\cellcolor{ForestGreen!20}$\mathbf{0.58}\,\pm\,0.05$} & \multicolumn{2}{c}{$6.34\,\pm\,0.56$} & \multicolumn{2}{c}{\cellcolor{ForestGreen!20}$\mathbf{1.50}\,\pm\,0.29$} & \multicolumn{2}{c}{\cellcolor{ForestGreen!20}$\mathbf{3.02}\,\pm\,0.33$} & \multicolumn{2}{c}{\cellcolor{ForestGreen!20}$\mathbf{1.35}\,\pm\,0.22$} & \multicolumn{2}{c}{\cellcolor{ForestGreen!20}$\mathbf{6.61}\,\pm\,0.99$} & \multicolumn{2}{c}{\cellcolor{ForestGreen!20}$\mathbf{2.08}\,\pm\,0.26$} & \multicolumn{2}{c}{$0.98\,\pm\,0.57$} & \multicolumn{2}{c}{$0.14\,\pm\,0.12$} & \multicolumn{2}{c}{$877.48\,\pm\,76.81$} \\
\texttt{Diffusion} & \multicolumn{2}{|c|}{$0.06\,\pm\,0.10$} & \multicolumn{2}{c}{$0.13\,\pm\,0.05$} & \multicolumn{2}{c}{$0.34\,\pm\,0.03$} & \multicolumn{2}{c}{$0.08\,\pm\,0.01$} & \multicolumn{2}{c}{$4.57\,\pm\,0.23$} & \multicolumn{2}{c}{$7.27\,\pm\,0.24$} & \multicolumn{2}{c}{$4.58\,\pm\,0.47$} & \multicolumn{2}{c}{$24.32\,\pm\,1.55$} & \multicolumn{2}{c}{$18.74\,\pm\,0.83$} & \multicolumn{2}{c}{$106.54\,\pm\,4.95$} & \multicolumn{2}{c}{$18.54\,\pm\,1.23$} & \multicolumn{2}{c}{$2.04\,\pm\,0.66$} & \multicolumn{2}{c}{$0.10\,\pm\,0.11$} & \multicolumn{2}{c}{$978.98\,\pm\,29.57$} \\
\texttt{GR00T} & \multicolumn{2}{|c|}{$0.04\,\pm\,0.09$} & \multicolumn{2}{c}{$0.05\,\pm\,0.03$} & \multicolumn{2}{c}{$0.50\,\pm\,0.02$} & \multicolumn{2}{c}{$0.12\,\pm\,0.02$} & \multicolumn{2}{c}{$4.70\,\pm\,0.35$} & \multicolumn{2}{c}{$12.16\,\pm\,0.63$} & \multicolumn{2}{c}{$2.56\,\pm\,0.50$} & \multicolumn{2}{c}{$32.41\,\pm\,1.85$} & \multicolumn{2}{c}{$41.87\,\pm\,2.31$} & \multicolumn{2}{c}{$233.11\,\pm\,6.53$} & \multicolumn{2}{c}{$22.17\,\pm\,1.39$} & \multicolumn{2}{c}{$23.60\,\pm\,2.49$} & \multicolumn{2}{c}{\cellcolor{ForestGreen!20}$\mathbf{0.00}\,\pm\,0.07$} & \multicolumn{2}{c}{$965.66\,\pm\,47.12$} \\
\texttt{Pi\_0.5} & \multicolumn{2}{|c|}{\cellcolor{ForestGreen!20}$\mathbf{0.40}\,\pm\,0.14$} & \multicolumn{2}{c}{\cellcolor{ForestGreen!20}$\mathbf{0.44}\,\pm\,0.07$} & \multicolumn{2}{c}{$0.34\,\pm\,0.02$} & \multicolumn{2}{c}{\cellcolor{ForestGreen!20}$\mathbf{0.05}\,\pm\,0.01$} & \multicolumn{2}{c}{$2.03\,\pm\,0.31$} & \multicolumn{2}{c}{\cellcolor{ForestGreen!20}$\mathbf{5.87}\,\pm\,0.88$} & \multicolumn{2}{c}{$2.16\,\pm\,0.59$} & \multicolumn{2}{c}{$10.44\,\pm\,1.76$} & \multicolumn{2}{c}{$11.37\,\pm\,0.75$} & \multicolumn{2}{c}{$57.57\,\pm\,4.17$} & \multicolumn{2}{c}{$7.60\,\pm\,1.29$} & \multicolumn{2}{c}{$1.94\,\pm\,1.38$} & \multicolumn{2}{c}{$0.12\,\pm\,0.12$} & \multicolumn{2}{c}{\cellcolor{ForestGreen!20}$\mathbf{710.78}\,\pm\,108.67$} \\
\texttt{Pi\_0.5 (LoRA)} & \multicolumn{2}{|c|}{$0.12\,\pm\,0.12$} & \multicolumn{2}{c}{$0.30\,\pm\,0.06$} & \multicolumn{2}{c}{$0.37\,\pm\,0.03$} & \multicolumn{2}{c}{$0.06\,\pm\,0.01$} & \multicolumn{2}{c}{$4.28\,\pm\,0.51$} & \multicolumn{2}{c}{$7.42\,\pm\,0.60$} & \multicolumn{2}{c}{$4.86\,\pm\,0.66$} & \multicolumn{2}{c}{$22.44\,\pm\,2.88$} & \multicolumn{2}{c}{$24.34\,\pm\,2.22$} & \multicolumn{2}{c}{$130.96\,\pm\,11.87$} & \multicolumn{2}{c}{$15.25\,\pm\,1.80$} & \multicolumn{2}{c}{\cellcolor{ForestGreen!20}$\mathbf{0.64}\,\pm\,0.50$} & \multicolumn{2}{c}{$0.22\,\pm\,0.13$} & \multicolumn{2}{c}{$895.10\,\pm\,72.36$} \\
\texttt{xVLA} & \multicolumn{2}{|c|}{$0.00\,\pm\,0.07$} & \multicolumn{2}{c}{$0.01\,\pm\,0.02$} & \multicolumn{2}{c}{$0.54\,\pm\,0.05$} & \multicolumn{2}{c}{$0.17\,\pm\,0.03$} & \multicolumn{2}{c}{$4.34\,\pm\,0.46$} & \multicolumn{2}{c}{$8.74\,\pm\,0.36$} & \multicolumn{2}{c}{$3.22\,\pm\,0.59$} & \multicolumn{2}{c}{$27.38\,\pm\,2.32$} & \multicolumn{2}{c}{$22.73\,\pm\,2.43$} & \multicolumn{2}{c}{$115.21\,\pm\,11.41$} & \multicolumn{2}{c}{$19.44\,\pm\,1.86$} & \multicolumn{2}{c}{$9.34\,\pm\,2.24$} & \multicolumn{2}{c}{\cellcolor{ForestGreen!20}$\mathbf{0.00}\,\pm\,0.07$} & \multicolumn{2}{c}{$974.08\,\pm\,29.35$} \\
\midrule
\texttt{Human} & \multicolumn{2}{|c|}{$1.00\,\pm\,0.00$} & \multicolumn{2}{c}{$1.00\,\pm\,0.00$} & \multicolumn{2}{c}{$0.40\,\pm\,0.03$} & \multicolumn{2}{c}{$0.05\,\pm\,0.00$} & \multicolumn{2}{c}{$0.42\,\pm\,0.04$} & \multicolumn{2}{c}{$0.58\,\pm\,0.06$} & \multicolumn{2}{c}{$0.24\,\pm\,0.11$} & \multicolumn{2}{c}{$1.56\,\pm\,0.17$} & \multicolumn{2}{c}{$23.24\,\pm\,1.28$} & \multicolumn{2}{c}{$95.97\,\pm\,5.04$} & \multicolumn{2}{c}{$1.11\,\pm\,0.14$} & \multicolumn{2}{c}{$0.00\,\pm\,0.05$} & \multicolumn{2}{c}{$0.00\,\pm\,0.05$} & \multicolumn{2}{c}{$83.78\,\pm\,9.01$} \\
\bottomrule
\end{tabular}
\end{adjustbox}
\label{tab:metrics_summary_combined}
\vspace{2pt}
\noindent\begin{minipage}{\linewidth}
\scriptsize
\textbf{Abbreviations:} Success = Success Rate [\%]; TP = Task Progression [\%]; BAVD = Bimanual Arm Velocity Difference [m/s]; BGVD = Bimanual Gripper Vertical Difference [m]; CPL = Cartesian Path Length [m]; CT = Completion Time [s]; ECC = Environment Collision Count; JPL = Joint Path Length [rad]; MCJ = Mean Cartesian Jerk [m/s\textsuperscript{3}]; MJJ = Mean Joint Jerk [rad/s\textsuperscript{3}]; OPL = Orientation Path Length [rad]; SCC = Self-Collision Count; SC = Slip Count; TL = Trajectory Length [steps].
\end{minipage}
\vspace{-1.5em}
\end{table*}

%% file: figures/significance_matrix.tex
\begin{figure*}[t!]
  \centering

  \begin{minipage}[b]{0.48\textwidth}
    \centering
    \includegraphics[width=\linewidth]{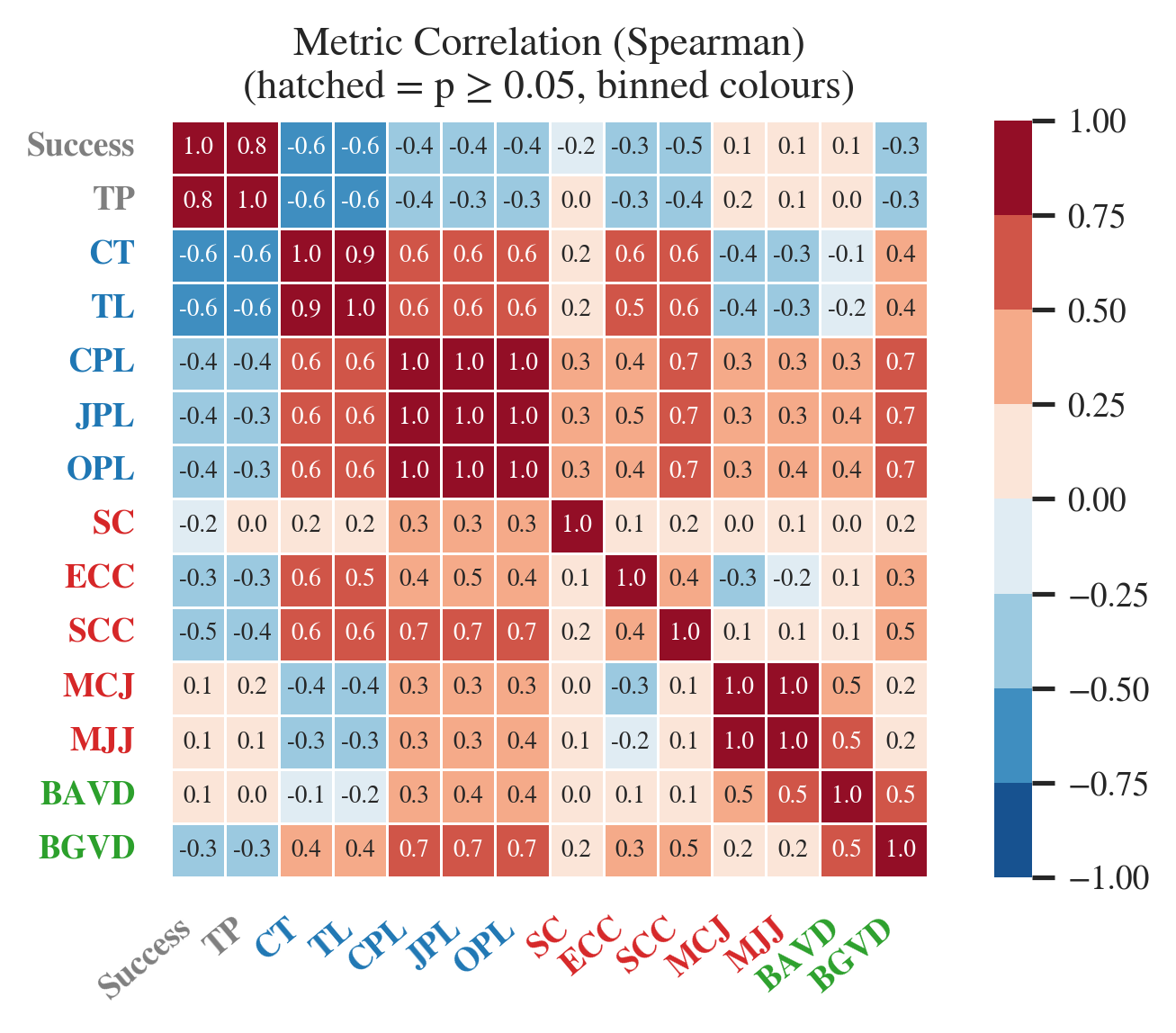}
  \end{minipage}
  \hfill
  \begin{minipage}[b]{0.48\textwidth}
    \centering
    \includegraphics[width=\linewidth]{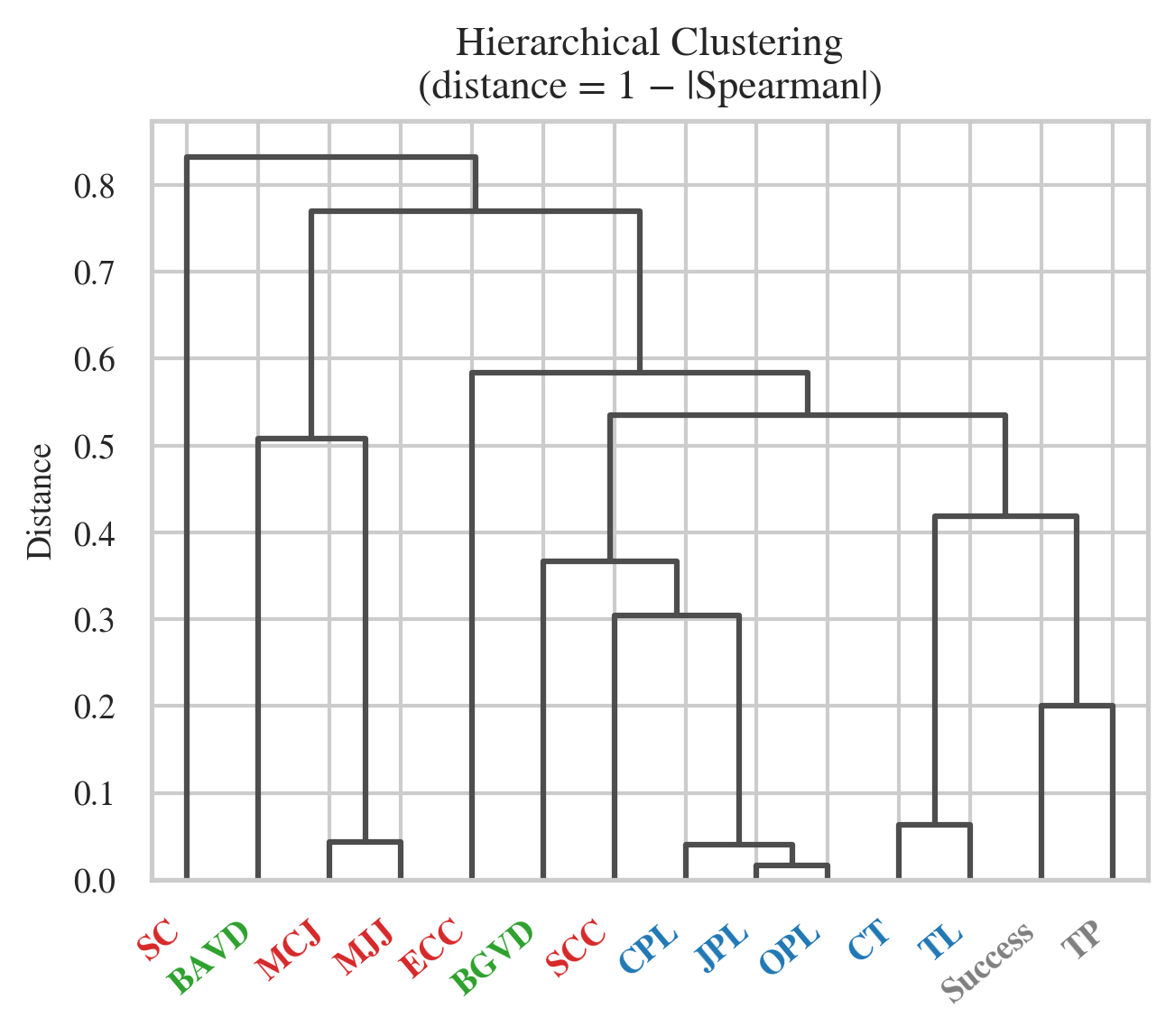}
  \end{minipage}

  \caption{\textbf{Correlation Structure of Behavioral Metrics.} From left to right,
(a)~Spearman correlation heatmap across all tasks and models. Hatched cells indicate non-significant correlations ($p \geq 0.05$). Color-coded labels denote metric groups: \textcolor{gray}{Outcome}, \textcolor{blue}{Efficiency}, \textcolor{red}{Safety/Stability}, and \textcolor{green}{Coordination}.
(b)~Average-linkage hierarchical clustering with distance $1 - |\rho_{\text{Spearman}}|$. The empirical clustering largely aligns with our predefined groupings, while confirming the metric suite is not collapsed into a single latent factor.}
  \label{fig:metric_dendrogram}
  \vspace{-1em}
\end{figure*}





%% file: figures/success_breakdown.tex
\begin{figure*}[htbp]
  \centering

  \begin{minipage}[b]{0.24\textwidth}
    \centering
    \includegraphics[width=\linewidth]{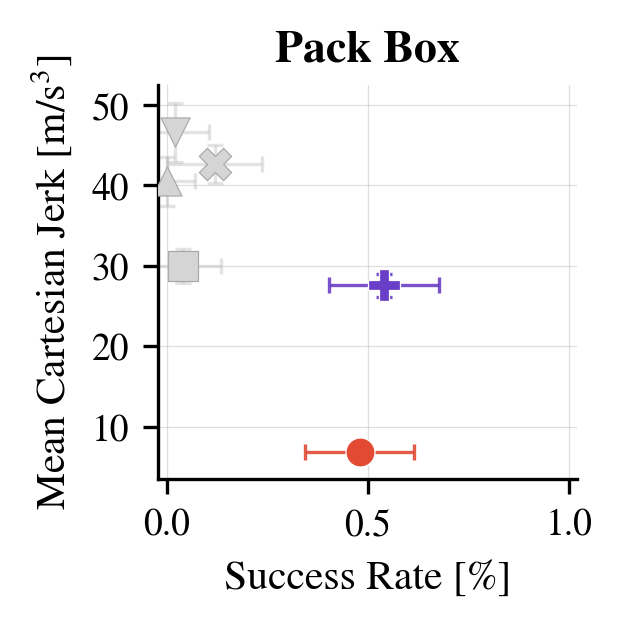}
  \end{minipage}%
  \hspace{0.005\textwidth}
  \begin{minipage}[b]{0.24\textwidth}
    \centering
    \includegraphics[width=\linewidth]{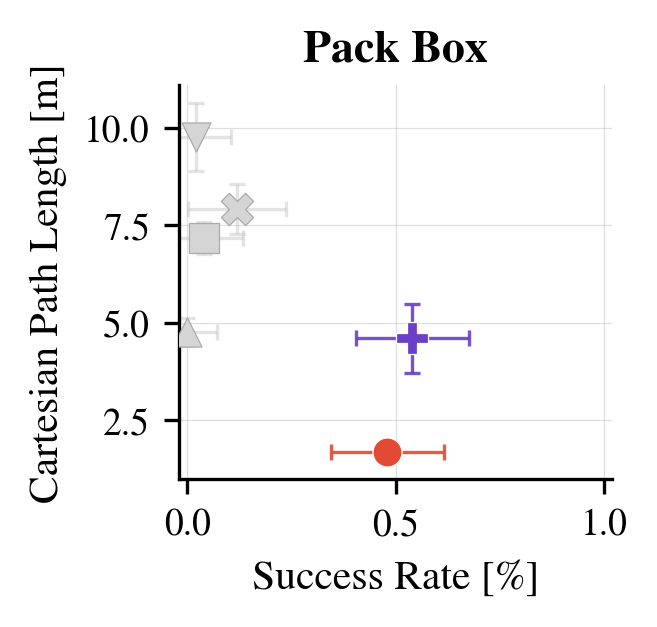}
  \end{minipage}%
  \hspace{0.005\textwidth}
  \begin{minipage}[b]{0.48\textwidth}
    \centering
    \includegraphics[width=\linewidth]{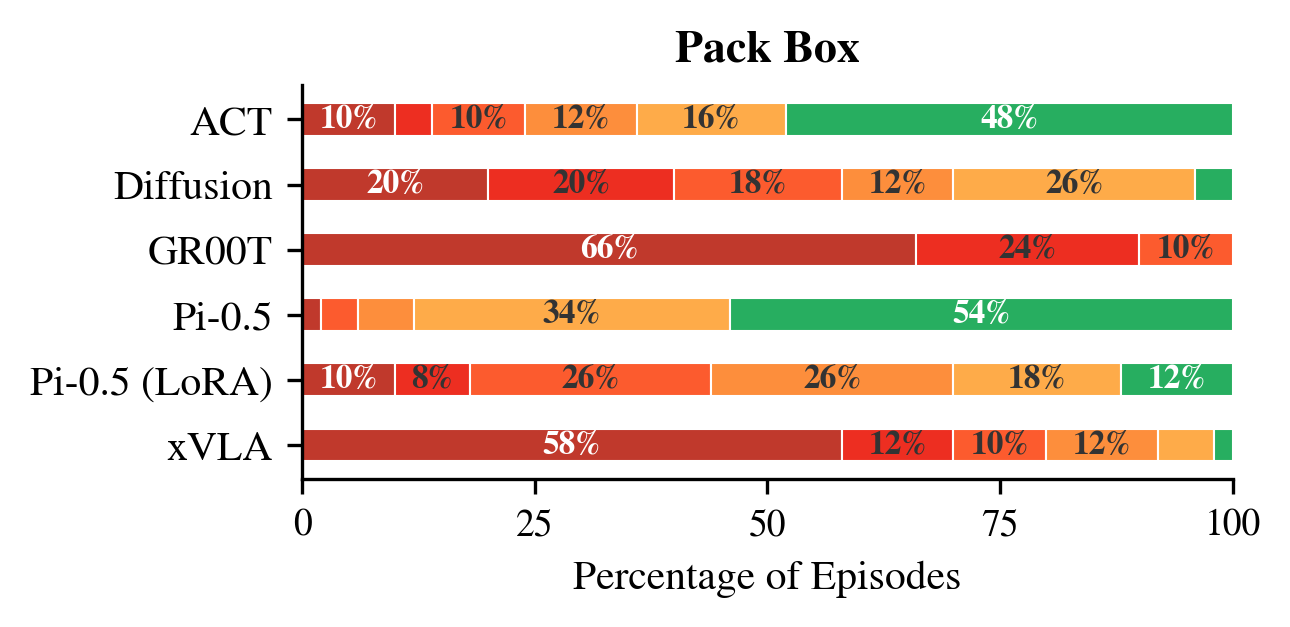}
  \end{minipage}

  \vspace{0.3em}
  \includegraphics[width=0.48\textwidth]{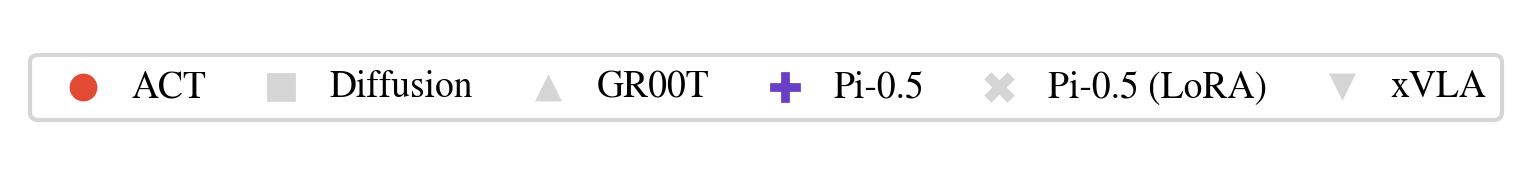}%
  \hspace{0.01\textwidth}
  \includegraphics[width=0.48\textwidth]{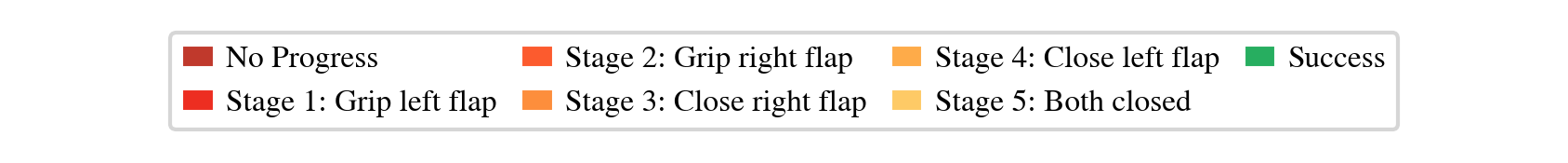}

  \caption{\textbf{Behavioral metrics and failure modes for \texttt{Pack Box}.} From left to right,
(a)~Mean Cartesian jerk vs.\ success rate and
(b)~Cartesian path length vs.\ success rate,
where each point represents a policy with 95\% confidence intervals.
\textcolor[HTML]{E24A33}{ACT} and \textcolor[HTML]{6A3FC7}{Pi-0.5} are highlighted;
Pi-0.5 achieves the highest success rate,
while ACT exhibits the lowest jerk and shortest path length,
indicating smoother and more efficient motion despite slightly lower success.
(c)~Failure mode breakdown showing the farthest subtask stage reached
by each policy's episodes.
Pi-0.5 succeeds in 54\% of trials and ACT in 48\%,
while GR00T and xVLA fail predominantly at early grasping stages.
Together, the scatter plots and failure analysis show that high success
does not guarantee efficient motion:
ACT's lower jerk and shorter paths suggest qualitatively different
execution strategies even among the top-performing policies.}

  \label{fig:packbox_diagnosis}
\end{figure*}

%% file: figures/variation_difference.tex
\begin{figure}[htbp]
  \centering

  \begin{minipage}[b]{\linewidth}
    \centering
    \includegraphics[width=0.8\linewidth]{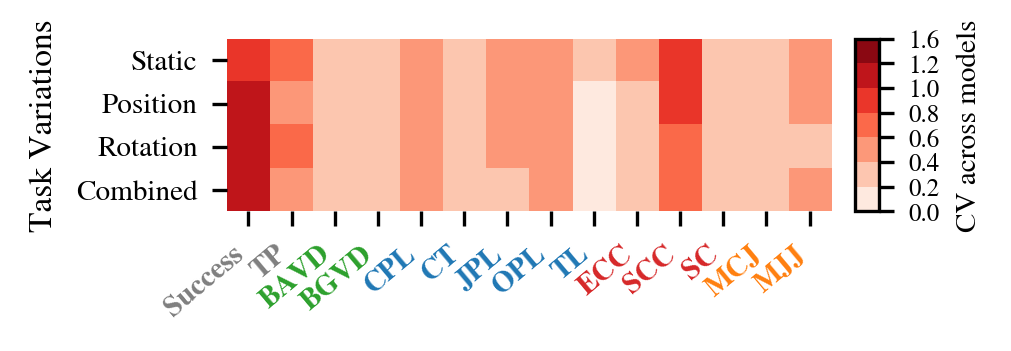}
    \vspace{-1em}
  \end{minipage}

  \begin{minipage}[b]{\linewidth}
    \centering
    \includegraphics[width=0.8\linewidth]{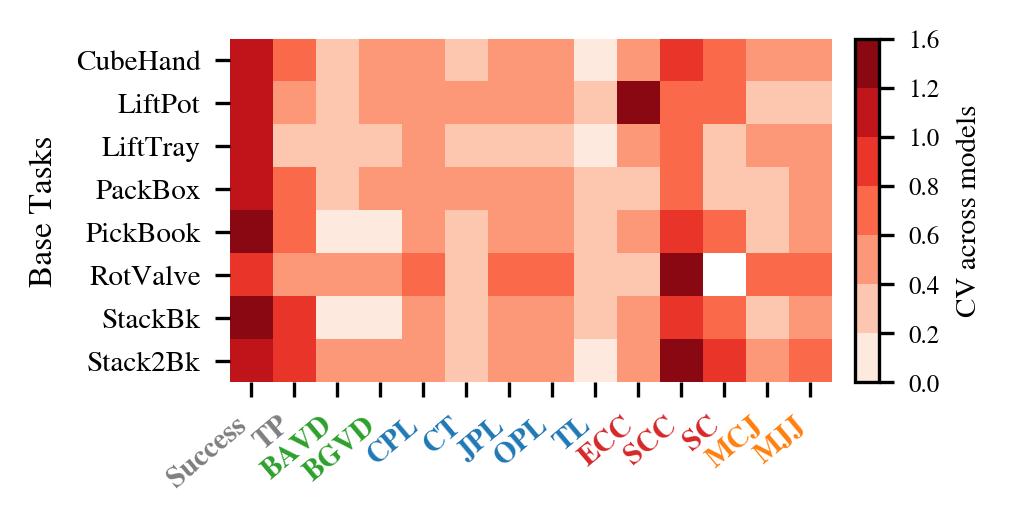}
  \end{minipage}

  \captionsetup[figure]{skip=1pt}
  \caption{\textbf{Coefficient of variation (CV) of metrics across policies, grouped by task variation (top) and base task (bottom).} Higher CV indicates greater discriminative power between policies.
  (Top)~Metric discrimination is largely stable across variation complexity.
  (Bottom)~Discrimination patterns are task-dependent: SCC is especially discriminating on \texttt{RotateValve} and \texttt{StackTwoBlocks}, while end-effector collision count (ECC) spikes on \texttt{LiftPot}. Success rate remains the most discriminating metric across all tasks.}
  \vspace{-1em}
  \label{fig:variation_diff}
\end{figure}

%% file: sections/6_discussion_conclusion.tex


\section{Discussion}
\benchmarkname~demonstrates that binary success alone is an incomplete measure of policy performance. By integrating behavioral metrics of fluency, precision, and coordination with outcome metrics that track stagewise progress and failure modes, the benchmark enables systematic analysis of how policies succeed, where they fail, and which execution factors underlie robustness. Our experiments show that these metrics are not only predictive of success but also expose distinctions hidden by success rates, validating the premise that richer evaluation reveals the true capabilities of a policy. Looking ahead, potential extensions include expanding variation modalities, and using the metrics as auxiliary objectives or reward-shaping signals. In this way, \benchmarkname~positions metrics as both a diagnostic lens and a training signal, advancing the development of interpretable, robust, and data-efficient robot learning.